\documentclass{article}

\usepackage{microtype}
\usepackage{graphicx}
\usepackage{subfigure}
\usepackage{booktabs} 

\usepackage{hyperref}

\usepackage[accepted]{icml2025}

\usepackage{amsmath}
\usepackage{amssymb}
\usepackage{mathtools}
\usepackage{amsthm}

\usepackage[capitalize,noabbrev]{cleveref}

\theoremstyle{plain}

\theoremstyle{definition}

\theoremstyle{remark}

\usepackage[textsize=tiny]{todonotes}

\usepackage{bbm}
\usepackage{dsfont}
\usepackage{amssymb}
\usepackage{pifont}
\usepackage{arydshln} 
\usepackage{makecell}

\newcommand{\cmark}{\ding{51}}%
\newcommand{\xmark}{\ding{55}}%

\usepackage[acronym]{glossaries}
\glsdisablehyper

\usepackage{titletoc}

\usepackage{placeins}

\newacronym{imp}{IMP}{Iterative Magnitude Pruning}
\newacronym{peft}{PEFT}{Parameter-Efficient Fine-Tuning}
\newacronym{per}{PERP}{Parameter-Efficient Retraining after Pruning}
\newacronym{ft}{FT}{Fine-Tuning}
\newacronym{lora}{LoRA}{Low-Rank Adaptation}
\newacronym{llms}{LLMs}{Large Language Models}
\newacronym{gpt}{GPT}{Generative Pretrained Transformer}
\newacronym{nlp}{NLP}{Natural Language Processing}
\newacronym{bn}{BN}{Batch-Normalization}
\newacronym{ln}{LN}{Layer-Normalization}
\newacronym{multlora}{ScaleLoRA}{ScaleLoRA}
\newacronym{masklora}{MaskLoRA}{MaskLoRA}
\newacronym{lora-prune}{LoRA-Prune}{LoRA-Prune}

\newcommand{\glsshort}[1]{\glsentryshort{#1}}

\icmltitlerunning{PERP: Rethinking the Prune-Retrain Paradigm in the Era of LLMs}

\begin{document}

\twocolumn[
\icmltitle{PERP: Rethinking the Prune-Retrain Paradigm in the Era of LLMs}



\icmlsetsymbol{equal}{*}

\begin{icmlauthorlist}
    \icmlauthor{\centering \textbf{Max Zimmer \quad Megi Andoni \quad Christoph Spiegel \quad Sebastian Pokutta}
    \\
    \small{Department for AI in Society, Science, and Technology, Zuse Institute Berlin, Germany\\
    Institute of Mathematics, Technische Universität Berlin, Germany}\\
    \texttt{\{zimmer,andoni,spiegel,pokutta\}@zib.de}}{}
\end{icmlauthorlist}

\vskip 0.3in
]




\begin{abstract}
    Neural Networks can be effectively compressed through \emph{pruning}, significantly reducing storage and compute demands while maintaining predictive performance. Simple yet effective methods like magnitude pruning remove less important parameters and typically require a costly retraining procedure to restore performance. However, with the rise of \glsentryshort{llms}, full retraining has become infeasible due to memory and compute constraints. This study challenges the practice of retraining all parameters by showing that updating a small subset of highly expressive parameters can suffice to recover or even enhance performance after pruning. Surprisingly, retraining just 0.01\%-0.05\% of the parameters in GPT-architectures can match the performance of full retraining across various sparsity levels, significantly reducing compute and memory requirements, and enabling retraining of models with up to 30 billion parameters on a \emph{single} GPU in minutes. To bridge the gap to full retraining in the high sparsity regime, we introduce two novel \glsentryshort{lora} variants that, unlike standard \glsentryshort{lora}, allow merging adapters back without compromising sparsity. Going a step further, we show that these methods can be applied for memory-efficient layer-wise reconstruction, significantly enhancing state-of-the-art retraining-free methods like Wanda \citep{Sun2023} and SparseGPT \citep{Frantar2023a}. Our findings present a promising alternative to avoiding retraining.
\end{abstract}

\section{Introduction}\emph{Pruning} \citep{Han2015, Gale2019, Lin2020, Hoefler2021, Zimmer2022} is among the state-of-the-art techniques to reduce the compute and storage requirements of Neural Networks, allowing to benefit from the extensive over-parametrization of modern architectures \citep{Zhang2016} throughout training while maintaining high performance with lower resource demands during deployment. Arguably simple yet effective approaches to obtaining such \emph{sparse} models follow the \emph{prune after training} paradigm and are exemplified by \glsfirst{imp} \citep{Han2015}, which starts from a pretrained \emph{dense} model and iteratively removes seemingly unimportant parameters followed by retraining to compensate for pruning-induced performance degradation.

\begin{figure}
    \centering
    \includegraphics[width=0.75\columnwidth]{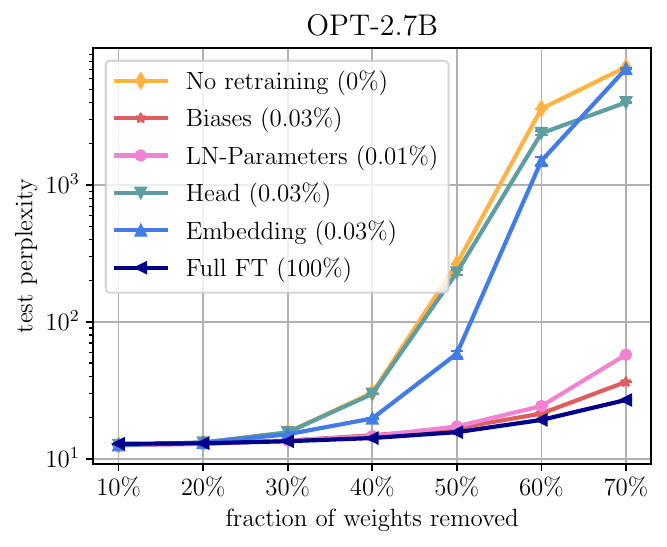}

    \caption{OPT-2.7B evaluated on WikiText: Final perplexity vs. sparsity after pruning, followed by retraining only the specified parameter subset. We indicate the percentage of trainable parameters in parentheses. Full \glsshort{ft} refers to full retraining of all parameters.}
    \label{fig:perplexity_vs_sparsity}
\end{figure}

Despite their popularity, approaches like IMP suffer from being computationally expensive, potentially having to perform many prune-retrain cycles and retraining epochs to obtain well-performing models for the task at hand. Especially given the surge in popularity of \emph{foundation models} -- huge pretrained models which are reused and fine-tuned for specific tasks -- a procedure such as IMP can be prohibitive in resource constrained environments \citep{Frantar2023a}, even when performing just a single prune-retrain cycle (\emph{One Shot}). Retraining hence enjoys a particularly negative reputation and a variety of approaches try to avoid it entirely, including novel weight-selection criteria for pruning without the need for retraining \citep{Frantar2023a, Sun2023, Theus2024, Jaiswal2023a}.

On the other hand, several works have tried to address the issue from the angle of making retraining itself less undesirable. \citet{Zimmer2021} accelerate retraining using a pruning-adaptive learning rate schedule, effectively reducing the number of iterations required while improving generalization performance. To find \emph{lottery tickets} \citep{Frankle2018} more efficiently, \citet{You2019} and \citet{Wolfe2021} try to find the pruning mask earlier in training, \citet{Jaiswal2023} speed up the mask-generation process by superimposing a set of masks throughout retraining, and \citet{Zhang2021b} reduce the number of retraining iterations by using only a critical subset of the data. \citet{Zimmer2023} show that constructing \emph{sparse model soups} during each phase of IMP can enhance its performance and the overall wall-time required for retraining.

In this work, we propose viewing the problem from another, previously underexplored angle, namely that of \emph{parameter-efficiency}. In almost all classical pruning literature, retraining after pruning is defined as a retraining of \emph{all} parameters at hand\footnote{For unstructured pruning, pruned parameters are forced to zero during retraining but are still part of backpropagation.}. This is particularly problematic with optimizers like Adam \citep{Kingma2014}, which need storage for parameters, gradients, and both first and second-order moments. With growing model sizes, especially in the context of \gls{llms}, retraining all parameters is a challenge both in terms of computational efficiency and storage demands. Notable exceptions to the retrain-all-parameters paradigm are recent works that introduce additional retraining parameters through \gls{lora} \citep{Hu2021b}. However, this comes with increased inference costs, as \gls{lora}-adapters usually cannot be merged with the original parameters without reducing model sparsity \citep{Sun2023}.

Our work is motivated by the following observation: Pruning can substantially degrade the model's performance, often to near-random levels. Yet, full retraining can restore performance in much fewer iterations than similar loss reductions would need during training from scratch \citep{Zimmer2021}, suggesting that retrained models retain considerable feature information and stay closely aligned with their pruned-but-not-retrained counterparts. The goal of this work is to recover these features efficiently by challenging the common practice of retraining all parameters after pruning. To that end, we investigate and propose \gls{peft} techniques \citep{Lialin2023a, Lialin2023} in the context of retraining after pruning, demonstrating that retraining a small subset of highly expressive parameters can effectively restore performance. Further, we propose two novel \gls{lora} variants that allow adapters to be merged back into the model without compromising sparsity.

Surprisingly, by retraining as little as 1\% of the parameters of state-of-the-art \gls{gpt} architectures such as OPT \citep{Zhang2022b}, LLaMA-2 \citep{Touvron2023}, Mistral \citep{Jiang2023} and Mixtral \citep{MistralMixtralExperts}, we recover nearly all of the full-retraining performance, even at moderate to high sparsity levels, where pruning without retraining collapses entirely. Similarly, retraining less than 0.5\% of the parameters of a ResNet-50 on ImageNet is sufficient to recover accuracy. We refer to the proposed approaches using the umbrella term \gls{per}. These methods enable pruning and retraining of up to 30 billion parameter \glspl{gpt} on a \emph{single} NVIDIA A100 GPU within minutes. Notably, they achieve this without increasing inference costs during deployment, as no additional parameters remain after retraining and the sparsity is preserved.

\paragraph{Contributions.} Our main contributions are:
\begin{enumerate}   
    \item \textbf{Restoring feature quality with few parameters.} We challenge the practice of retraining all parameters after pruning, demonstrating that retraining a small subset of highly expressive parameters such as biases or \gls{ln}-parameters can effectively restore performance after pruning, with backpropagation of less than 0.05\% of the total parameters often sufficing for full recovery. Despite initial performance degradation to near-random levels, retraining a fraction of the parameters can restore qualitative features with significantly reduced compute and memory requirements, allowing retraining of up to 30 billion parameters models on a \emph{single} NVIDIA A100 GPU within minutes.    
    \item \textbf{Closing the gap: Low-Rank Adaptation.} We propose and discuss two novel \gls{lora}-based methods that enable efficient retraining while preserving model sparsity. Unlike standard \gls{lora}, our approaches allow merging adapters back into the original weights without compromising sparsity. Using as little as 1\% of parameters, we achieve performance comparable to or better than full retraining across various sparsity levels and model architectures.
    
    \item \textbf{Efficient Layer-wise Reconstruction.} Going a step further, we demonstrate that such parameter-efficient approaches can significantly enhance state-of-the-art retraining-free methods like SparseGPT and Wanda when applied in a layer-wise reconstruction setting. By optimizing each layer independently with further reduced memory overhead, we improve zero-shot accuracy by up to 17\% for magnitude pruning and up to 4\% for Wanda and SparseGPT.

\end{enumerate}

Our work challenges the practice of retraining all parameters after pruning, highlighting that the features retained in pruned models can be effectively realigned to good performance by retraining a remarkably small subset of parameters, such as only the \gls{ln}. We emphasize that our goal is not to propose a set of methods that claim state-of-the-art performance, but to systematically explore parameter-efficient retraining of sparse models from the ground up.

\section{Methodology and Experimental Setup}\label{sec:methodology}
We begin with a short overview of pruning and \gls{peft}.
\subsection{Preliminaries}\label{subsec:preliminaries}

\textbf{Pruning.}
We prune \gls{llms} in a post-hoc fashion, removing individual or groups of weights of a pretrained model, following the \emph{prune after training} paradigm. Our focus lies on the \emph{One Shot} setting, where the network is pruned once and then retrained to recover the pruning-induced performance drop, as opposed to iterative approaches such as IMP. Since pruning a non-trivial portion of the parameters may result in significant performance degradation, the retraining step is fundamental to obtain a well-performing model, which is typically possible in much fewer iterations than what standard training would require to achieve a comparable reduction in train loss \citep{Zimmer2021}. In this work, we study pruning to produce high-quality sparse models rather than to obtain lottery tickets \citep{Frankle2018}.

Arguably the simplest and most popular pruning criterion is \emph{magnitude pruning}, which removes the fraction of weights with smallest magnitude. However, it is by far not the only one, as detailed in studies like \citet{LeCun1989, Hassibi1992, Molchanov2016, Yeom2019}, see \citet{Hoefler2021} for a comprehensive review. Despite its simplicity, magnitude pruning and its variants have been shown to be highly effective for convolutional architectures, being able to produce sparse models comparable in performance to those from much more complex algorithms \citep{Gale2019, Zimmer2021}. 

For \gls{llms}, magnitude pruning has been argued to be unsuited, with perplexity exploding already at moderate sparsity levels, and it being no better than random pruning at higher sparsity \citep{Yin2023a}. In consequence, there is growing interest in developing pruning criteria other than magnitude that yield high-performance models without the need for retraining \citep{Kwon2022, Frantar2023a, Sun2023}. Both \citet{Sun2023} and \citet{Yin2023a} explain the inability to magnitude-prune \gls{llms} with observations made by \citet{Dettmers2022} regarding the \emph{emergence of large magnitude features} in transformers beyond a certain size. These large features, a small yet significant subset of hidden features, are critical for model performance, and pruning them severely impacts predictive accuracy \citep{Sun2023}; a problem that magnitude pruning fails to address. \citet{Yin2023a} take these outliers into account and propose a mechanism to better allocate sparsity between the layers. Interestingly, \citet{Yin2023a} note that the previous efficacy of magnitude pruning is closely intertwined with the feasibility of retraining, which is in turn considered infeasible for models of the scale of \gls{llms}.

\emph{Layer-wise reconstruction} offers an efficient alternative to full retraining by breaking down the problem into per-layer subproblems. As opposed to optimizing all parameters with a \emph{global} loss, such approaches use a calibration dataset on which a \emph{local} per-layer reconstruction error is minimized. Specifically, for layer $l$ with input $X_l$, weights $W_l$, and binary pruning mask $M_l$, the goal is to optimize
\begin{equation}\label{eq:reconstruction}
    \min_{\hat W_l}\Vert W_l X_l - (M_l \odot \hat W_l) X_l\Vert_2^2,
\end{equation}
where $\hat W_l$ represents the reconstructed weights, and $\odot$ denotes the element-wise Hadamard product.

\textbf{Transfer learning.}
As models grow in size, \gls{ft} -- the process of adapting a pretrained model to a novel task -- has become the norm, avoiding the inefficiencies of training from scratch for each new task \citep{Houlsby2019, Kumar2022}. \gls{ft} capitalizes on the transfer of existing knowledge to a closely related domain (\emph{transfer learning}). Yet, the immense size of foundation models can make the \gls{ft} approach more challenging, requiring storage for the entire model, its gradients, and auxiliary buffers, even for brief training. In response, various \glsfirst{peft} methods have emerged. They significantly reduce the number of trainable parameters, cutting down on compute and storage needs, while preserving performance levels comparable to conventional full \gls{ft}.

\gls{peft} methods are broadly categorized as selective, additive, or reparametrization-based \citep{Lialin2023a}. \emph{Selective methods} update specific model components, such as the top linear layer \citep{Kumar2022a, Evci2022a}, only the biases \citep{Zaken2021}, or by partitioning specific tensors into active and inactive portions \citep{Vucetic2022}. \emph{Additive methods}, like \emph{adapters} \citep{Houlsby2019, He2022a}, add new parameters which are trained for specific tasks while the main model remains unchanged. \emph{Reparametrization-based methods} exploit the low intrinsic dimensionality of fine-tuning \citep{Aghajanyan2020}. A well-known example is \gls{lora} \citep{Hu2021b}, which implicitly enforces low-rank constraints on additive updates to pretrained parameter matrices, substantially decreasing the number of trainable parameters. Precisely, \gls{lora} freezes the pretrained parameters and reparametrizes each matrix $W_0 \in \mathbb{R}^{n \times m}$ as $W_0 + \Delta W$, where $\Delta W = BA$ represents the update. In this representation, $B \in \mathbb{R}^{n \times r}$ and $A \in \mathbb{R}^{r \times m}$ implicitly constrain the rank of $\Delta W$ to be at most $r \ll \min(n,m)$. $B$ is zero-initialized, thereby preserving the original model's behavior. During training, only $A$ and $B$ are updated, while $W_0$ remains fixed.

\textbf{Other related literature.} \citet{Kwon2022} propose a structured pruning framework for transformers, explicitly avoiding retraining for efficiency. \citet{Zhang2023a} develop a training-free pruning method inspired by prune-and-grow strategies from \emph{Dynamic Sparse Training} \citep{Evci2019}. Several works propose techniques in the domain of sparse fine-tuning in transfer learning. \citet{Zhang2023} address the problem of performing gradient-based pruning by utilizing the \gls{lora} gradients as substitute for the full gradient. \citet{Liu2021d} aim at pruning pretrained models for improvements when fine-tuning to downstream tasks. \citet{Li2022} reduce the number of parameters for weight importance computation in sparse fine-tuning. While conventional retraining typically involves retraining all parameters, \citet{Sun2023} further fine-tune pruned \gls{llms} using \gls{lora}, however at the price of increased inference costs since the \gls{lora} matrices cannot be merged with the original parameters. Similarly, \citet{Ma2023} perform \gls{lora} after structured pruning, eventually allowing \gls{lora} matrices to be merged. Concurrent to our work, \citet{Lu2024} and \citet{Munoz2024} propose \gls{peft} strategies that maintain sparsity. \citet{Lu2024} focus on retraining multiplicative rescaling factors of pretrained parameters, while \citet{Munoz2024} integrate masking of \gls{lora} matrices with quantization and neural architecture search. We contrast their approaches with ours in \cref{subsec:lora}. To the best of our knowledge, our work is the first to thoroughly investigate \gls{peft} for retraining and reconstruction after pruning, presenting several efficient methods that recover performance without increasing inference costs or reducing sparsity.

\subsection{Parameter-Efficient Retraining After Pruning}
Pruning can degrade the model's performance to near-random levels. Yet, retraining often restores performance in much fewer iterations than similar loss reductions during pretraining \citep{Zimmer2021}. This optimization often involves merely a few iterations, even when dealing with substantial pruning-induced performance degradation. Consequently, even if the pruned network is severely damaged, it likely retains most of the task-informative features. \textit{We hypothesize that, similar to fine-tuning in transfer learning, retraining can be significantly more efficient by leveraging these features rather than adjusting the entire network, despite pruning severely damaging the model.} In essence, we view pruning as a disturbance to the features the model had learned and retraining after pruning is about refining imperfect, yet valuable features.

\textbf{What gains can we expect from parameter-efficiency?}
Parameter-Efficient Retraining aims to substantially reduce the computational load and memory demands of backpropagation by retraining fewer parameters, i.e., freezing the majority of parameters to not require gradients. While techniques like adapters or LoRA might slightly increase computational requirements in the forward pass, we expect that the computational load is still significantly reduced. However, a major benefit also lies in the significant reduction in memory requirements. Typically, optimizers such as AdamW \citep{Kingma2014, Loshchilov2017} require multiple buffers per parameter and retraining fewer parameters results in considerably less allocated memory. This reduction is crucial for retraining large models efficiently, exemplified by our ability to retrain up to 30 billion parameter models on a single NVIDIA A100. Often, the memory required for storing activations during backpropagation can be significantly reduced. However, if parameters early in the network require gradients, then almost all activations need to be stored for backpropagation. To further reduce these requirements, we investigate the use of parameter-efficient retraining in the layer-wise reconstruction setting as per \autoref{eq:reconstruction}, limiting the optimization to individual layers.

\subsection{Experimental setup}\label{subsec:experimental_setup}
We outline our general experimental approach, detailing datasets, architectures, and metrics. To enable reproducibility, our code is available at \href{https://github.com/ZIB-IOL/PERP}{github.com/ZIB-IOL/PERP}.

Our study primarily investigates language modeling in \glsshort{nlp}. We use pretrained \gls{gpt} models available through HuggingFace \citep{Wolf2020}, namely \emph{OPT-1.3B/2.7B/6.7B/13B/30B} \citep{Zhang2022b}, \emph{LLaMA-2-7B/13B} \citep{Touvron2023}, \emph{Mistral-7B} \citep{Jiang2023} as well as \emph{Mixtral-8x7B} \citep{MistralMixtralExperts}. Retraining is done on the \emph{C4} dataset \citep{Raffel2020a} with context-length sequence sizes using AdamW \citep{Loshchilov2017} using a linear schedule and a 10\% warmup. For validation, we randomly sample 100 sequences from the validation split. The models are evaluated using the perplexity metric on the \emph{WikiText} dataset \citep{Merity2016}. In addition, we provide the zero-shot accuracy on the EleutherAI evaluation set \citep{Gao2023}. We follow \citet{Sun2023} and prune all linear layers except the embedding and final linear head, assigning uniform sparsity to all layers and providing experiments for unstructured and the semi-structured 2:4 and 4:8 sparsity patterns \citep{Mishra2021}. In \autoref{app:conv_results}, we also provide experiments for image classification.

\section{Parameter-Efficient Retraining}\label{sec:per}
Pruning can be seen as distorting the initially acquired features, diminishing the network's expressivity by settling on suboptimal features. With most parameters set to be immutable, our goal is to regain performance (minimizing perplexity or maximizing accuracy) with a minimal number of parameters. To that end, we first examine subsets of the parameters with varying complexity, which we hypothesize to hold significant expressive power during retraining.
\subsection{Restoring feature quality with few parameters}\label{subsec:selective}
We investigate the following parameter subsets:

\textbf{Biases}: We only retrain the network's biases. Despite corresponding to only a small fraction of the total parameters, biases can be crucial for model expressivity; \citet{Zaken2021} specifically propose a \gls{peft} method that only adjusts these biases to the new task. 

\textbf{\glsentryfirst{ln}}: \gls{ln} layers include trainable scaling and bias parameters. The importance of normalization layers for expressivity has been highlighted by \citet{Mudrakarta2018, Giannou2023}, with \citet{Frankle2020b} demonstrating that training only these parameters can enable otherwise frozen, randomly-initialized networks to achieve significant accuracy. 

\textbf{Linear Head}: A popular \gls{peft} approach is \emph{Linear Probing}, where all parameters remain fixed except for the final linear head to align the features to the new task. 

\textbf{Embedding}: The embedding layer is the network's first layer and is crucial for processing input data.

\autoref{fig:perplexity_vs_sparsity} shows the (log) perplexity after magnitude pruning OPT-2.7B followed by retraining only the specified parameter subset, indicating the percentage of trainable parameters in parentheses. All approaches are retrained for 1000 iterations, tuning the initial learning rate, see \autoref{app:technical_details} for full details. We make the following observations:

\begin{enumerate}
    \item \textbf{Pruning without retraining collapses.} Consistent with previous findings \citep[e.g.,][]{Yin2023a}, magnitude pruning without subsequent retraining results in a collapse even at moderate sparsity levels.
    \item \textbf{Head and embedding retraining are ineffective.} Contrary to our expectations, retraining only the head or the embedding layer fails to recover performance lost due to pruning. Although there are slight improvements over not retraining, the impact seems marginal.
    \item \textbf{Biases and \gls{ln}-parameters restore performance.} Surprisingly, retraining biases or \gls{ln} recovers most of the lost performance, despite corresponding to just 0.03\% and 0.01\% of the total parameters, respectively. Even at high sparsity levels, both approaches nearly match the perplexity of full \gls{ft}, with significantly reduced memory and compute requirements.
\end{enumerate}

\autoref{tab:perplexity_and_accuracy_oneshot_comparison} shows the final perplexity (upper halves) and average zero-shot accuracy (lower halves) after magnitude pruning OPT-2.7B and OPT-30B to 30\%-70\% unstructured sparsity and retraining. Retraining only biases and \gls{ln}-parameters effectively restores much of the performance lost to pruning, particularly when measuring zero-shot accuracy. This supports our hypothesis: Although pruning disrupts learned features, adjusting per-layer translations via bias retraining or the affine \gls{ln} transformations is sufficient to achieve well-performing sparse networks. However, a gap remains at high sparsity levels like 70\%. Note that, due to memory constraints, OPT-2.7B is the largest model that can be fully retrained on a single GPU, so full \gls{ft} for the 30 billion parameter variant is omitted, since multiple GPUs would be required. In contrast, optimizing only biases or \gls{ln}-parameters for the 30B model is feasible on a single GPU, demonstrating the memory efficiency of parameter-efficient retraining. We further note that retraining the dense (non-sparse) model on C4 offers no benefits, and these results transfer to the other architectures, as detailed in \autoref{app:additional_experiments}.

\begin{table}[t]
    \caption{OPT-2.7B/30B: Parameter-efficient approaches vs. full retraining with 30\%-70\% of the parameters pruned. The first column lists the method, and the second shows the percentage of trainable parameters. Full \gls{ft} represents the standard retraining baseline and is only possible for OPT-2.7B due to memory constraints. The next five columns display the average mean perplexity (upper halves, lower is better) and average zero-shot accuracy (lower halves, higher is better) across multiple seeds, with standard deviations excluded for clarity. 
    \\}
    \label{tab:perplexity_and_accuracy_oneshot_comparison}
    \centering
    \resizebox{\columnwidth}{!}{%
    \begin{tabular}{lc ccccc}
    \large{\textbf{OPT-2.7B}}\\
    \toprule
    \footnotesize{\textbf{Perplexity: 12.47}} & & \multicolumn{5}{c}{\textbf{Sparsity}}\\
    \cmidrule{3-7}
        \textbf{Method} & \% trainable & 30\% & 40\% & 50\% & 60\% & 70\% \\
    \midrule
    Full \gls{ft} & 100\% & 13.42 & 14.16 & 15.63 & 19.20 & 26.86  \\
    \midrule
    \glsshort{masklora} & 0.882\% & 13.41 & 14.24 & 15.75 & 18.37 & 25.58  \\
    Biases & 0.034\% & 13.57 & 14.86 & 16.56 & 21.41 & 37.24  \\
    \gls{ln}-Parameters & 0.013\% & 13.58 & 14.63 & 17.24 & 24.26 & 58.08  \\
    No retraining & 0.000\% & 15.58 & 30.31 & 265.13 & 3604.96 & 7252.22  \\
    
    \midrule
    \footnotesize{\textbf{Accuracy: 47.81\%}} & & \multicolumn{5}{c}{\textbf{Sparsity}}\\
    \cmidrule{3-7}
        \textbf{Method} & \% trainable & 30\% & 40\% & 50\% & 60\% & 70\% \\
    \midrule 
    Full \gls{ft} & 100\% & 46.99\% & 46.20\% & 45.44\% & 44.53\% & 42.44\%  \\
    \midrule
    \glsshort{masklora} & 0.882\% & 47.25\% & 46.29\% & 45.92\% & 43.92\% & 41.56\%  \\
    Biases & 0.034\% & 46.75\% & 45.66\% & 45.29\% & 42.75\% & 39.49\%  \\
    \gls{ln}-Parameters & 0.013\% & 46.78\% & 45.48\% & 44.72\% & 41.37\% & 38.32\%  \\
    No retraining & 0.000\% & 44.99\% & 42.77\% & 40.01\% & 35.34\% & 32.38\%  \\
    \bottomrule
    
    \\
    \large{\textbf{OPT-30B}}\\
    \toprule
    \footnotesize{\textbf{Perplexity: 9.55}} & & \multicolumn{5}{c}{\textbf{Sparsity}}\\
    \cmidrule{3-7}
        \textbf{Method} & \% trainable & 30\% & 40\% & 50\% & 60\% & 70\% \\
    \midrule
    \glsshort{masklora} & 0.329\% & 10.27 & 11.04 & 11.75 & 13.55 & 16.65  \\
    Biases & 0.013\% & 10.32 & 11.30 & 12.58 & 14.66 & 20.06  \\
    \gls{ln}-Parameters & 0.005\% & 10.29 & 11.17 & 12.50 & 15.17 & 21.41  \\
    No retraining & 0.000\% & 12.37 & 24.29 & 168.06 & 11676.00 & 28180.15  \\
    
    \midrule
    \footnotesize{\textbf{Accuracy: 55.07\%}} & & \multicolumn{5}{c}{\textbf{Sparsity}}\\
    \cmidrule{3-7}
        \textbf{Method} & \% trainable & 30\% & 40\% & 50\% & 60\% & 70\% \\
    \midrule
    \glsshort{masklora} & 0.329\% & 54.42\% & 53.71\% & 53.12\% & 50.60\% & 47.79\%  \\
    Biases & 0.013\% & 53.84\% & 52.90\% & 51.59\% & 50.00\% & 46.33\%  \\
    \gls{ln}-Parameters & 0.005\% & 53.31\% & 52.39\% & 51.88\% & 49.93\% & 45.07\%  \\
    No retraining & 0.000\% & 51.57\% & 44.19\% & 36.39\% & 32.01\% & 31.92\%  \\

    \bottomrule
    
    \end{tabular}
    }
\end{table}

\subsection{Closing the gap: Low-Rank Adaptation}\label{subsec:lora}
We demonstrated that suitable parameters-subsets are often sufficient to restore performance, despite updating only a small fraction of the parameters. Especially in the low to medium sparsity regime, retraining only biases and \gls{ln}-parameters poses a compute- and memory-efficient alternative to full \gls{ft}. However, in the higher sparsity regime, full retraining prevails. Combining parameter subsets, such as retraining both biases and \gls{ln}-parameters simultaneously, yields improvements. However, the effect is less pronounced, indicating diminishing returns from adding more such parameter subsets to align the distorted features (cf.\ \autoref{app:ablations} for a full ablation). To close this gap, we investigate the usage of reparametrization-based approaches to efficiently update \emph{all} parameters.

Previous works \citep{Sun2023, Ma2023} propose to use \gls{lora} for retraining after pruning, demonstrating that well-performing sparse models can be achieved even with restricted, low-dimensional reparametrizations. Yet, adapting \gls{lora} to the prune-retrain paradigm poses challenges. For dense models, \gls{lora} does not increase inference costs during deployment since eventually undoing the reparametrization by setting $W \gets W + BA$ and then removing $B$ and $A$ recovers the original architecture, and the forward pass can be evaluated using just $W$. However, for pruning, integrating the dense matrix $BA$ compromises the sparsity of the pruned tensor $W$. Precisely, in unstructured weight pruning, the matrix $W$ has an irregular pattern of zeros; instead of merging $BA$ into $W$ after retraining, which would destroy this sparsity, $B$, $A$ and $W$ need to be retained as separate matrices in the model, effectively increasing storage and compute requirements during inference.

An immediate fix to this issue is to apply the binary sparsity mask $M$ of $W$ to $BA$ before merging, i.e., setting $W \gets W + M \odot BA$ at the end of retraining, where $\odot$ again denotes element-wise Hadamard multiplication. However, this approach, which we term \glsentryshort{lora-prune}, results in a noticeable increase in perplexity because pruning the update matrix $BA$ again disrupts the model. To avoid this problem, we propose two new \gls{lora} methods that are compatible with pruning. These methods enable efficient and effective retraining of pruned models without compromising sparsity, allowing adapters to be merged back into the model. The key idea is adapting the reparametrization, which modifies the forward pass of standard \gls{lora}. Recall that in \gls{lora}, the forward pass is computed as $Wx + B(A(\text{dropout}(x)))$ for a sample $x$ (biases are omitted for clarity).

\paragraph{\glsentryshort{multlora}:} To maintain the sparsity of $W$ when eventually merging $BA$ into $W$, a natural choice is to learn multiplicative adapters instead of additive ones. Specifically, we redefine the \gls{lora} forward pass as $\left [(BA)\odot W\right ]x$. While zeros in $W$ remain zero after incorporating adapters through $W \gets (BA)\odot W$, this approach also introduces challenges. Dropout cannot be applied directly, though we found this to have minimal impact. Unlike classical \gls{lora}, which leverages matrix multiplication associativity to compute $B(Ax)$ efficiently, \glsentryshort{multlora} requires explicit construction of $BA$, adding slight memory overhead. Further, standard \gls{lora} maintains the original model's forward pass by initializing $B=0$ and $A$ randomly, ensuring $BA=0$ before retraining. However, for \glsentryshort{multlora}, this would require $BA = \mathds{1}_{n, m}$ for $B\in \mathbb{R}^{n, r}$ and $A\in \mathbb{R}^{r, m}$, which we resolve by initializing $B=\mathds{1}_{n, r}/\sqrt{r}$ and $A=\mathds{1}_{r, m}/\sqrt{r}$. Although this initialization can slightly hinder optimization, we empirically found that it performs best among all tested variants that preserve $(BA) \odot W = W$ before retraining. Alternatively, reparametrizing the forward pass as $\left [(\mathds{1}_{n, m} + BA)\odot W\right ]x$ with $B=0$ and $A$ randomly-initialized is possible, but less effective. We note that the latter approach shares some similarity with the one by \citet{Lu2024} (cf.\ \autoref{app:technical_details}).

\paragraph{\glsentryshort{masklora}:} This method preserves sparsity without altering the additive nature of \gls{lora}, unlike \glsentryshort{multlora}. It masks pruned elements of $BA$ during the forward pass, processing sample $x$ as $(W + M \odot BA)x$. After retraining, the original architecture is restored by updating $W \gets W + M \odot BA$ and removing matrices $B$ and $A$. In contrast to \glsentryshort{lora-prune}, \glsentryshort{masklora} incorporates the sparsity of $W$ into the training process. This approach prevents the tendency of classical \gls{lora} to train $B$ and $A$ in a way that negates the zeros in $W$, which can lead to a significant drop in performance when re-pruning before merging the parameters. By integrating the sparsity pattern, \glsentryshort{masklora} avoids this issue, resulting in no performance degradation upon merging. Although \glsentryshort{masklora} can incorporate dropout, we found no significant performance improvement. Again, the Hadamard product requires computing the full matrix $BA$ in the forward pass, causing minor memory overhead. Unlike \glsentryshort{multlora}, \glsentryshort{masklora} allows for the default \gls{lora} initialization with $B=0$. Since $M$ is applied in every iteration, it must be either inferred during the forward pass or cached, the latter being impractical for large models due to memory constraints. The computational overhead of deriving $M$ from $W$ each iteration can be mitigated, as we discuss at the end of this section. Concurrent to our work, \citet{Munoz2024} also propose a \gls{lora}-variant that masks the product $BA$ for eventual merging.

\autoref{tab:lora_acc_comparison} presents a comparison of the zero-shot accuracies across all presented \gls{lora}-variants tested on \gls{gpt} models under unstructured pruning (50\% sparsity) and semi-structured 2:4 and 4:8 sparsities. The second column shows whether the adapters can be merged back without compromising sparsity, being the case for all but standard \gls{lora}. We reparametrize all linear layers except the embedding, and further also retrain biases and \gls{ln}-parameters. We tune the initial learning rate for each variant, with iterations fixed at 1000, \gls{lora} rank $r$ at 16, and rescaling $\alpha$ at 32 (see \Cref{app:technical_details} for details). The trainable parameter fraction for all \gls{lora}-variants ranges between 0.48\% and 0.95\% of the full model, varying by model. Our conclusions are:

\begin{enumerate}
    \item \textbf{\glsshort{lora-prune} degrades performance:} Pruning $BA$ at the end, as done in \glsentryshort{lora-prune}, reduces performance compared to \gls{lora}, which in turn cannot be merged after retraining.
    \item \textbf{\glsshort{multlora} and \glsshort{masklora} are effective:} Both methods underperform slightly compared to \gls{lora}, but they nearly match its performance and sometimes even outperform it, despite allowing parameter merging after training without compromising sparsity.
    \item \textbf{\glsshort{masklora} closes the gap to full \gls{ft}:} \autoref{tab:perplexity_and_accuracy_oneshot_comparison} shows how \glsentryshort{masklora} reduces the gap between previous methods and full \gls{ft}, often surpassing it while utilizing only a fraction as little as 0.5\% of the parameters.
\end{enumerate}

\begin{table}
    \caption{OPT-13B/30B, LLaMA-2-7B/13B, and Mixtral-8x7B: EleutherAI zero-shot accuracy comparison of all \gls{lora}-variants retraining magnitude pruned models, both in the unstructured pruning setting (50\% sparsity), as well as for the semi-structured 2:4 and 4:8 sparsities. The second column indicates whether the adapters can be merged into the original weights without destroying the sparsity. 
    We report the mean accuracy over several seeds and omit the standard deviation for the sake of clarity.\\ }
    \label{tab:lora_acc_comparison}
    \centering
    \resizebox{\columnwidth}{!}{%
    \begin{tabular}{lcc  cc | cc | c}
    \toprule
     & & & \multicolumn{2}{c|}{\textbf{OPT}} & \multicolumn{2}{c|}{\textbf{LLaMA-2}} & \multicolumn{1}{c}{\textbf{Mixtral}}\\
    \cmidrule{4-8}
        \textbf{Method} & Mergeable & Sparsity & 13B & 30B & 7B & 13B & 8x7B \\
    \midrule
    Baseline & -- & 0\% & 52.60\% & 55.07\% & 59.69\% & 62.99\% & 67.70\%\\
    \midrule
    \glsentryshort{lora} & \xmark & 50\% & 50.88\% & 52.65\% & 56.14\% & 59.21\% & 65.33\%  \\
    \glsentryshort{lora-prune} & \cmark & 50\% & 47.40\% & 50.86\% & 55.98\% & 58.56\% & 64.17\%  \\
    \textbf{\glsentryshort{multlora}} & \cmark & 50\% & 50.50\% & 52.31\% & 55.79\% & 59.45\% & 64.84\%  \\
    \textbf{\glsentryshort{masklora}} & \cmark & 50\% & 49.65\% & 52.97\% & 55.67\% & 59.44\% & 64.84\%  \\
    \midrule
    \glsentryshort{lora} & \xmark & 2:4 & 49.93\% & 49.67\% & 51.32\% & 55.68\% & 60.69\%  \\
    \glsentryshort{lora-prune} & \cmark & 2:4 & 48.63\% & 50.09\% & 51.17\% & 54.56\% & 59.25\%  \\
    \textbf{\glsentryshort{multlora}} & \cmark & 2:4 & 49.73\% & 49.77\% & 51.15\% & 55.24\% & 60.84\%  \\
    \textbf{\glsentryshort{masklora}} & \cmark & 2:4 & 50.00\% & 49.97\% & 51.41\% & 55.73\% & 61.01\%  \\
    \midrule
    \glsentryshort{lora} & \xmark & 4:8 & 50.49\% & 51.01\% & 54.13\% & 58.30\% & 62.78\%  \\
    \glsentryshort{lora-prune} & \cmark & 4:8 & 48.96\% & 49.52\% & 53.34\% & 57.36\% & 60.72\%  \\
    \textbf{\glsentryshort{multlora}} & \cmark & 4:8 & 50.11\% & 50.88\% & 53.50\% & 57.07\% & 62.50\%  \\
    \textbf{\glsentryshort{masklora}} & \cmark & 4:8 & 50.66\% & 50.90\% & 54.29\% & 57.99\% & 62.72\%  \\
    \bottomrule
    \end{tabular}
    }
    \end{table}

    \begin{table*}
        \caption{OPT-30B - Retraining: Task performance improvement by retraining with \glsentryshort{masklora} for magnitude pruning, Wanda, and SparseGPT, displaying 60\% unstructured sparsity. Each entry shows the improvement compared to no retraining, e.g., +0.5\% indicates a 0.5\% performance increase. We report the mean performance over several seeds and omit the standard deviation for the sake of clarity.\\ }
        \label{tab:task_accuracy_comparison_high_sparsity_retraining}
        \centering
        \resizebox{\textwidth}{!}{%
        \begin{tabular}{lc rrrrrrr|r}
        \toprule
        \textbf{Method} & Sparsity & \multicolumn{8}{c}{$\Delta$ Task Accuracy} \\
        \cmidrule{3-10}
        & & BoolQ & RTE & HSwag & WinoG & ARC-e & ARC-c & OBQA & Average \\
        \midrule
        Magnitude & 60\% & +28.07\% & +1.44\% & +22.93\% & +14.64\% & +37.98\% & +12.24\% & +14.40\% & +18.82\%  \\
        Wanda & 60\% & +1.56\% & +4.69\% & +6.05\% & +3.95\% & +6.99\% & +3.28\% & +3.20\% & +3.84\%  \\
        SparseGPT & 60\% & +4.48\% & +0.90\% & +1.57\% & +0.32\% & +0.55\% & +0.55\% & +2.10\% & +0.99\%  \\
    
        \bottomrule
        \end{tabular}
        }
    \end{table*}

\begin{figure}
        \centering
        \includegraphics[width=0.75\columnwidth]{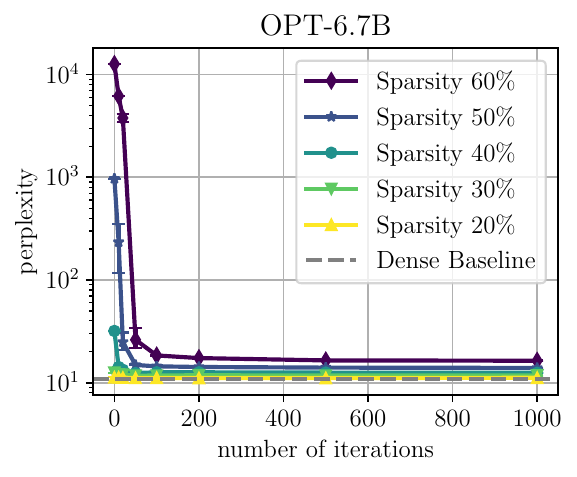}

        \caption{OPT-6.7B evaluated on WikiText: Final perplexity after retraining using \glsentryshort{masklora} for as many iterations as indicated on the x-axis. \glsentryshort{masklora} retrains roughly 1\% of the parameters.}
        \label{fig:perplexity_vs_iterations}
\end{figure}

\paragraph{Efficiency considerations.}
Before detailing the computational overhead of \glsentryshort{multlora} and \glsentryshort{masklora} compared to standard \gls{lora}, we note their efficiency and storage benefits over full \gls{ft}. We can retrain a 30B parameter model on a single NVIDIA A100 GPU, highlighting memory efficiency, whereas full retraining of OPT-30B requires multiple GPUs. Parameter-efficient retraining not only reduces storage costs but also increases efficiency. For example, on OPT-2.7B, full retraining achieves 3500 train tokens per second (tps), while \glsentryshort{multlora} and \glsentryshort{masklora} reach up to 5200 tps, and updating only biases and normalization parameters increases it to 8100 tps (cf.\ \autoref{tab:throughput_comparison}). As shown in \autoref{fig:perplexity_vs_iterations}, \glsshort{masklora} quickly reduces the perplexity of OPT-6.7B across various sparsity levels. Without retraining, perplexity rises from $10^1$ to $10^4$, but \glsshort{masklora} significantly lowers it, saturating after a few iterations.

The computational overhead of \glsentryshort{multlora} compared to \gls{lora} is minimal, with throughputs of 5200 tps and 5300 tps, respectively. Unmodified \glsentryshort{masklora} is slower at 3000 tps, but this can be improved by masking only every few iterations instead, though it slightly increases perplexity; we hence decided to not follow this path. Attempts to schedule masking frequency showed no benefit, and caching masks, while efficient, is impractical for larger models. Finally, finding no advantage in using dropout, we streamlined the process by adding matrices before the forward pass instead of performing forward for $W$ and $M \odot BA$ separately. Further optimization using a TorchScript-compiled function increased throughput to 4700 tps, nearly matching \gls{lora}.

\begin{table}
    \caption{OPT-2.7B: Retraining throughput comparison in tokens per second (tps) across different retraining methods.\\ }
    \label{tab:throughput_comparison}
    \centering
    \resizebox{\columnwidth}{!}{%
    \begin{tabular}{cccccc}
    \toprule
    \textbf{Full \gls{ft}} & \textbf{\gls{lora}} & \textbf{\glsshort{multlora}} & \textbf{\glsshort{masklora}} & \textbf{\glsshort{masklora}} & \textbf{Biases + \gls{ln}} \\
     &  &  & (standard) & (optimized) & \\
    \midrule
    3,500 & 5,300 & 5,200 & 3,000 & 4,700 & 8,100 \\
    \bottomrule
    \end{tabular}
    }
\end{table}

In summary, our results demonstrate that applying methods such as \glsshort{masklora} suffices to reach the performance achievable through full retraining, while requiring a minimal number of efficient iterations for performance recovery. 

\subsection{Efficient Layer-wise Reconstruction}

When using any of the previously introduced approaches, all activations of the network need to be stored for backpropagation. Despite retraining only a small fraction of parameters, these memory demands can still be prohibitive for large models. To further reduce the memory footprint, we investigate the use of \glsentryshort{masklora} in the layer-wise reconstruction setting as per \autoref{eq:reconstruction}. Precisely, we compare magnitude pruning, Wanda \citep{Sun2023}, and SparseGPT \citep{Frantar2023a}, each with and without solving the reconstruction problem after determining the mask $M_l$ through the respective methods. To that end, we reparametrize the reconstruction weights $\hat W_l$ using \glsshort{masklora} to obtain parameter-efficiency and preserve the sparsity pattern enforced by $M_l$. We optimize each layer-block sequentially using AdamW on 128 random samples from C4, with \autoref{eq:reconstruction} as the loss function.

As opposed to magnitude pruning, both SparseGPT and Wanda naturally rely on calibration data, the random selection of which influences the quality of the final result \citep{Williams2023}. We use the same set for both methods as well as the subsequent reconstruction. Note that we now only need to store a fraction of activations, buffers, and the computational graph for the parameters for each layer-block, rather than a fraction of the entire model. For OPT-30B, this amounts to roughly 0.35\% of a single layer.

As shown in \autoref{tab:accuracy_reconstruction_comparison}, using \glsentryshort{masklora} for reconstruction \emph{significantly enhances the performance of all three pruning methods}, including SparseGPT, which itself already performs a reconstruction step. The reconstruction optimization incurs minimal overhead when processing calibration data anyways, being the case for both Wanda and SparseGPT. With \glsentryshort{masklora} reconstruction, we observe improvements of up to 17\% in zero-shot accuracy for magnitude pruning and up to 4\% for Wanda and SparseGPT. Surprisingly, full \gls{ft} performs significantly worse than \glsentryshort{masklora} in the high sparsity regime (cf.\ \autoref{tab:fullft_reconstruction_accuracy_oneshot_comparison} in \autoref{app:additional_experiments}), likely due to overfitting on the calibration data.

Comparing magnitude pruning with reconstruction against Wanda without reconstruction, magnitude pruning often performs better than expected and can even outperform Wanda; we think this is a fair comparison since both methods require calibration data once. In general, magnitude pruning fails to address large features which are critical for performance, yet simple reconstruction with \glsentryshort{masklora} can significantly enhance it, sometimes even surpassing Wanda with reconstruction at high sparsities (cf.\ \autoref{app:ablations}). Reconstruction is more memory-efficient, but retraining (i.e., non layer-wise) performs better in high sparsity scenarios, in particular when evaluating individual tasks (cf.\ \autoref{tab:task_accuracy_comparison_high_sparsity_retraining}).

\begin{table}[h]
    \caption{OPT-2.7B/6.7B/13B/30B - Reconstruction: Zero-shot accuracy comparison on the EleutherAI evaluation set. We report magnitude pruning, Wanda and SparseGPT with and without \glsentryshort{masklora} when reconstructing, for unstructured 50\% sparsity and semi-structured 2:4 and 4:8 sparsities. 
    We report the mean accuracy over several seeds and omit the standard deviation for the sake of clarity.\\ }
    \label{tab:accuracy_reconstruction_comparison}
    \centering
    \resizebox{\columnwidth}{!}{%
    \begin{tabular}{lcc cccc}
    \toprule
     & & & \multicolumn{4}{c}{\textbf{OPT}}\\
    \cmidrule{4-7}
        \textbf{Method} & Reconstruction & Sparsity & 2.7B & 6.7B & 13B & 30B \\
    \midrule
    Baseline & -- & 0\% & 47.81\% & 51.55\% & 52.60\% & 55.07\%\\
    \midrule
    Magnitude & \xmark & 50\% & 40.07\% & 35.54\% & 33.80\% & 36.39\%  \\
    Magnitude & \cmark & 50\% & 45.14\% & 48.99\% & 50.41\% & 51.81\%  \\
    Wanda & \xmark & 50\% & 42.63\% & 47.14\% & 50.34\% & 53.15\%  \\
    Wanda & \cmark & 50\% & 46.47\% & 49.81\% & 51.65\% & 54.00\%  \\
    SparseGPT & \xmark & 50\% & 46.53\% & 50.26\% & 51.93\% & 54.01\%  \\
    SparseGPT & \cmark & 50\% & 46.62\% & 50.42\% & 51.92\% & 54.33\%  \\
    \midrule
    Magnitude & \xmark & 2:4 & 35.94\% & 36.38\% & 36.65\% & 34.92\%  \\
    Magnitude & \cmark & 2:4 & 43.54\% & 47.70\% & 49.16\% & 47.74\%  \\
    Wanda & \xmark & 2:4 & 42.82\% & 46.12\% & 47.70\% & 49.69\%  \\
    Wanda & \cmark & 2:4 & 43.69\% & 47.46\% & 48.56\% & 50.93\%  \\
    SparseGPT & \xmark & 2:4 & 44.12\% & 47.28\% & 48.84\% & 51.17\%  \\
    SparseGPT & \cmark & 2:4 & 44.67\% & 48.39\% & 50.38\% & 52.33\%  \\
    \midrule
    Magnitude & \xmark & 4:8 & 36.95\% & 36.90\% & 36.13\% & 36.81\%  \\
    Magnitude & \cmark & 4:8 & 44.59\% & 48.51\% & 50.19\% & 50.52\%  \\
    Wanda & \xmark & 4:8 & 44.02\% & 47.45\% & 49.10\% & 51.20\%  \\
    Wanda & \cmark & 4:8 & 44.69\% & 48.64\% & 50.21\% & 51.83\%  \\
    SparseGPT & \xmark & 4:8 & 44.98\% & 48.33\% & 50.12\% & 52.28\%  \\
    SparseGPT & \cmark & 4:8 & 45.99\% & 49.48\% & 50.96\% & 52.92\%  \\
    \bottomrule
    \end{tabular}
    }
    \end{table}

\section{Discussion}\label{sec:discussion}
We demonstrated that retraining a minimal fraction of parameters, such as biases or \gls{ln}-parameters, effectively mitigates pruning-induced performance drops. By backpropagating as little as 0.05\% of the parameters compared to full retraining, we significantly reduce compute and memory demands. We further proposed two efficient \gls{lora}-variants, allowing adapters to be merged back after retraining without compromising sparsity or increasing inference time. Applying these methods for layer-wise reconstruction further reduces memory demands while enhancing state-of-the-art retraining-free methods. Our findings make retraining after pruning a viable option for large models, and we hope to stimulate further research on both training-free pruning criteria and efficient retraining.

\newpage

\section*{Acknowledgements}
This research was partially supported by the DFG Cluster of Excellence MATH+ (EXC-2046/1, project id 390685689) funded by the Deutsche Forschungsgemeinschaft (DFG) as well as by the German Federal Ministry of Education and Research (fund number 01IS23025B).

\bibliography{references}
\bibliographystyle{icml2025}

\newpage
\appendix
\onecolumn

\FloatBarrier
\section{Technical details and training settings}
\FloatBarrier
\label{app:technical_details}

\FloatBarrier
\subsection{Pretraining settings and metrics.}
\FloatBarrier
For \gls{nlp} tasks, we use pretrained models from Huggingface and specify only the retraining settings as outlined in \Cref{subsec:experimental_setup}.

For computer vision, we focus on \emph{ImageNet} \citep{Russakovsky2015}, utilizing \emph{ResNet} architectures \citep{He2015} and measuring performance with top-1 test accuracy. We follow standard practices by retraining networks with momentum SGD, allocating 10\% of the training data for validation, and using the ALLR learning rate schedule \citep{Zimmer2021} for retraining. We follow \citet{Zimmer2021} and globally prune everything except biases and \gls{bn} parameters. As opposed to \gls{nlp}, we perform the pretraining process ourselves. \autoref{tab:problem_config} details our pretraining configurations, including the number of epochs, batch size, weight decay, and learning rate. We opt for SGD as the optimizer, though we recognize a range of other optimization methods are available \citep[see e.g.,][]{Kingma2014, Pokutta2020}. We maintain the default momentum value of 0.9. In the last column of the table we report the performance achieved with standard dense training, using top-1 test accuracy as the metric for image classification tasks, which denotes the percentage of test samples correctly classified.

\begin{table}[h]
\caption{Exact pretraining configurations in our vision experiments.}
\label{tab:problem_config}
\begin{center}
\resizebox{\textwidth}{!}{%
     \begin{tabular}{ll llllll}
\toprule
Dataset & Network (number of weights) & Epochs & Batch size & Weight decay & Learning rate ($t$ = training epoch) & Unpruned test accuracy \\
\midrule
ImageNet & ResNet-50 (26 M) & 90 & 256 & 1e-4 & linear from 0.1 to 0 & 76.12\% {\footnotesize \textpm 0.01\%} \\
\bottomrule
\end{tabular}
}
\end{center}
\end{table}

\FloatBarrier
\subsection{Pruning and Retraining}\label{app:prune_retrain}
\FloatBarrier

\paragraph{Pruning settings.}
Effective pruning relies on the accurate identification of weights to prune and the distribution of sparsity among the layers. \citet{Zhu2017} introduced the \textsc{Uniform} allocation, pruning each layer by the same relative amount. \citet{Gale2019} improved this with \textsc{Uniform+}, keeping the first convolutional layer dense and limiting pruning in the final fully-connected layer to 80\%. \citet{Evci2019} adapted the Erd\H{o}s-R\'enyi Kernel (ERK) \citep{Mocanu2018} for layerwise sparsity, accounting for layer dimensions. \citet{Lee2020a} proposed Layer-Adaptive Magnitude-based Pruning (LAMP), targeting minimal output distortion at pruning, assessed through $L_2$-distortion on worst-case inputs.

In \gls{nlp}, following \citet{Sun2023}, we prune all linear layers except embeddings and the final classification head, applying uniform sparsity throughout. For a comparison of diverse selection schemes for \gls{llms}, see \citet{Yin2023a}. Our experiments include both unstructured sparsity and semi-structured 2:4 and 4:8 sparsities. In vision tasks, aligning with \citet{Zimmer2021, Evci2019, Dettmers2019}, we prune everything except biases and \gls{bn} parameters, employing the \textsc{Global} criterion which treats all parameters as a single vector and computes a universal threshold for pruning.

\paragraph{Hyperparameters for Retraining: The Learning Rate.}
In computer vision, automating the learning rate schedule for retraining has received increased interest, aiming to circumvent the need for tuning the schedule in each phase. We describe various schedules where $T$ is the total number of epochs of original training with a learning rate schedule $(\eta_t)_{t\leq T}$, and $T_{{rt}}$ is the number of epochs in each retraining phase. FT \citep{Han2015} uses a constant learning rate, $\eta_T$, from the final epoch of initial training. LRW \citep{Renda2020} repeats the last $T - T_{{rt}}$ epochs of the original schedule. SLR \citep{Le2021} compresses the initial schedule into the retraining period with an initial warm-up. CLR \citep{Le2021} uses a cosine-based schedule with a warm-up to $\eta_1$. LLR \citep{Zimmer2021} linearly decays from $\eta_1$ to zero in each cycle. For vision tasks, we adopt ALLR \citet{Zimmer2021}, using a linear schedule that adjusts the initial rate based on the impact of pruning and available retraining time, balancing cycle length and pruning-induced performance degradation.

For \gls{llms}, we stick to AdamW with a linear learning rate decay from a tuned initial value. We experiment with starting values 5e-6, 1e-5, 5e-5, 1e-4 and 5e-4.

\paragraph{Hyperparameters for Retraining: Batch size and Weight decay.}
For vision, we retain the same batch size and weight decay parameters as used in pretraining. However, for \gls{llms} we set the weight decay to zero and found no improvement in increasing this value. We use a batch size of 2 and gradient accumulation for 4 steps for all models with less than 30 billion parameters. For larger models, we use a batch size of 1 and 2 gradient accumulation steps. We use gradient checkpointing to reduce the memory demands at the expense of efficiency.

\paragraph{Other sparsity-preserving \gls{peft} approaches.}
\glsshort{multlora} modifies the \gls{lora} forward pass to $\left [(BA)\odot W\right ]x$, requiring $B$ and $A$ to be initialized differently from classical \gls{lora}. An alternative is to reparametrize as $\left [(\mathds{1}_{n, m} + BA)\odot W\right ]x$, with $B=0$ and $A$ randomly initialized. This method is similar to \citet{Lu2024}, who use an outer product of matrices or vectors applied to each dimension of the pruned matrix $W$, but not directly through low-rank adapters as in \gls{lora}. While we experimented with the low-rank adapted variant of the method of \citet{Lu2024}, i.e., using the forward pass $\left [(\mathds{1}_{n, m} + BA)\odot W\right ]x$, we found it to not converge as fast as \glsshort{multlora}, which consistently performed better.

\newpage

\section{Full experiments}\label{app:additional_experiments}
In this section, we provide full results, following the same structure as \Cref{sec:per}.
\FloatBarrier
\subsection{Restoring feature quality with few parameters}
\FloatBarrier

\begin{figure}[h]
    \centering
    \includegraphics[width=0.5\columnwidth]{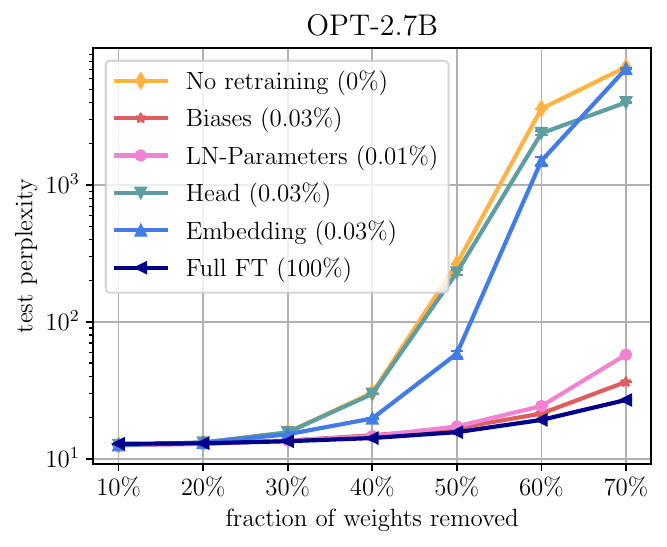}

    \caption{OPT-2.7B evaluated on WikiText: Final perplexity vs. sparsity after pruning, followed by retraining only the specified parameter subset. We indicate the percentage of trainable parameters in parentheses. Full \glsshort{ft} refers to full retraining of all parameters.}

\end{figure}

\begin{figure}[h]
    \centering
    \includegraphics[width=0.5\columnwidth]{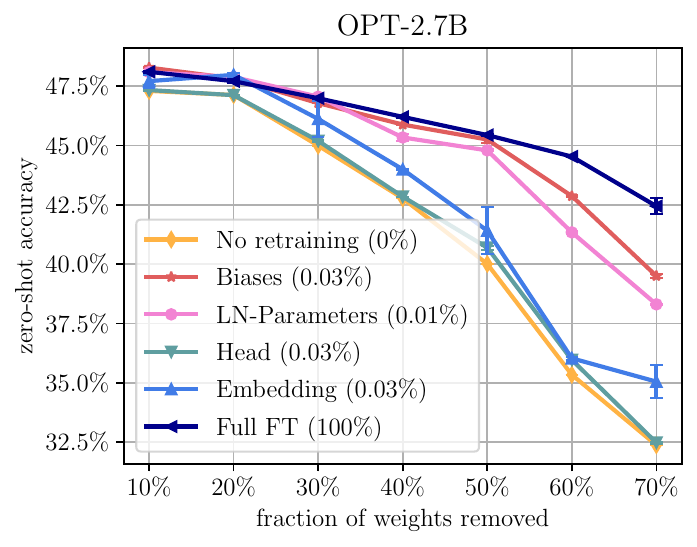}

    \caption{OPT-2.7B evaluated on the EleutherAI tasks: Final average zero-shot accuracy vs. sparsity after pruning, followed by retraining only the specified parameter subset. We indicate the percentage of trainable parameters in parentheses. Full \glsshort{ft} refers to full retraining of all parameters.}
    \label{fig:acc_vs_sparsity}
\end{figure}

\begin{table}[h]
    \caption{OPT-2.7B/30B: Parameter-efficient approaches vs. full retraining with 30\%-70\% of the parameters pruned. The first column lists the method, and the second shows the percentage of trainable parameters. Full \gls{ft} represents the standard retraining baseline and is only possible for OPT-2.7B due to memory constraints. The next five columns display the average mean perplexity (upper halves, lower is better) and average zero-shot accuracy (lower halves, higher is better) across multiple seeds, with standard deviations excluded for clarity. The dense models attain a perplexity and accuracy of 12.47 and 47.81\%, as well as 9.55 and 55.07\% for OPT-2.7B/30B, respectively.\\}
    \centering
    \begin{tabular}{lc ccccc}
    \large{\textbf{OPT-2.7B}}\\
    \toprule
    \footnotesize{\textbf{Perplexity: 12.47}} & & \multicolumn{5}{c}{\textbf{Sparsity}}\\
    \cmidrule{3-7}
        \textbf{Method} & \% trainable & 30\% & 40\% & 50\% & 60\% & 70\% \\
    \midrule
    Full \gls{ft} & 100\% & 13.42 & 14.16 & 15.63 & 19.20 & 26.86  \\
    \midrule
    \glsshort{masklora} & 0.882\% & 13.41 & 14.24 & 15.75 & 18.37 & 25.58  \\
    Biases & 0.034\% & 13.57 & 14.86 & 16.56 & 21.41 & 37.24  \\
    \gls{ln}-Parameters & 0.013\% & 13.58 & 14.63 & 17.24 & 24.26 & 58.08  \\
    No retraining & 0.000\% & 15.58 & 30.31 & 265.13 & 3604.96 & 7252.22  \\
    
    \midrule
    \footnotesize{\textbf{Accuracy: 47.81\%}} & & \multicolumn{5}{c}{\textbf{Sparsity}}\\
    \cmidrule{3-7}
        Method & \% trainable & 30\% & 40\% & 50\% & 60\% & 70\% \\
    \midrule 
    Full \gls{ft} & 100\% & 46.99\% & 46.20\% & 45.44\% & 44.53\% & 42.44\%  \\
    \midrule
    \glsshort{masklora} & 0.882\% & 47.25\% & 46.29\% & 45.92\% & 43.92\% & 41.56\%  \\
    Biases & 0.034\% & 46.75\% & 45.66\% & 45.29\% & 42.75\% & 39.49\%  \\
    \gls{ln}-Parameters & 0.013\% & 46.78\% & 45.48\% & 44.72\% & 41.37\% & 38.32\%  \\
    No retraining & 0.000\% & 44.99\% & 42.77\% & 40.01\% & 35.34\% & 32.38\%  \\
    \bottomrule
    
    \\
    \large{\textbf{OPT-30B}}\\
    \toprule
    \footnotesize{\textbf{Perplexity: 9.55}} & & \multicolumn{5}{c}{\textbf{Sparsity}}\\
    \cmidrule{3-7}
        \textbf{Method} & \% trainable & 30\% & 40\% & 50\% & 60\% & 70\% \\
    \midrule
    \glsshort{masklora} & 0.329\% & 10.27 & 11.04 & 11.75 & 13.55 & 16.65  \\
    Biases & 0.013\% & 10.32 & 11.30 & 12.58 & 14.66 & 20.06  \\
    \gls{ln}-Parameters & 0.005\% & 10.29 & 11.17 & 12.50 & 15.17 & 21.41  \\
    No retraining & 0.000\% & 12.37 & 24.29 & 168.06 & 11676.00 & 28180.15  \\
    
    \midrule
    \footnotesize{\textbf{Accuracy: 55.07\%}} & & \multicolumn{5}{c}{\textbf{Sparsity}}\\
    \cmidrule{3-7}
        Method & \% trainable & 30\% & 40\% & 50\% & 60\% & 70\% \\
    \midrule
    \glsshort{masklora} & 0.329\% & 54.42\% & 53.71\% & 53.12\% & 50.60\% & 47.79\%  \\
    Biases & 0.013\% & 53.84\% & 52.90\% & 51.59\% & 50.00\% & 46.33\%  \\
    \gls{ln}-Parameters & 0.005\% & 53.31\% & 52.39\% & 51.88\% & 49.93\% & 45.07\%  \\
    No retraining & 0.000\% & 51.57\% & 44.19\% & 36.39\% & 32.01\% & 31.92\%  \\

    \bottomrule
    
    \end{tabular}
\end{table}

\begin{table}[h]
    \caption{LLaMA-2-7B/13B: Parameter-efficient retraining approaches with 30\%-70\% of the parameters pruned. The first column lists the method, and the second shows the percentage of trainable parameters. The next five columns display the average mean perplexity (upper halves, lower is better) and average zero-shot accuracy (lower halves, higher is better) across multiple seeds, with standard deviations excluded for clarity. Note that both models do not have biases, retraining only these is hence not possible. \\}
    \label{apptab:perplexity_oneshot_comparison_llama_small}
    \centering
    \begin{tabular}{lc ccccc}
    \large{\textbf{LLaMA-2-7B}}\\
    \toprule
    \footnotesize{\textbf{Perplexity: 5.11}} & & \multicolumn{5}{c}{\textbf{Sparsity}}\\
    \cmidrule{3-7}
        \textbf{Method} & \% trainable & 30\% & 40\% & 50\% & 60\% & 70\% \\
    \midrule
    \glsshort{masklora} & 0.590\% & 5.38 & 5.74 & 6.49 & 8.04 & 11.75  \\
    \gls{ln}-Parameters & 0.004\% & 5.47 & 6.02 & 7.63 & 10.77 & 30.55  \\
    No retraining & 0.000\% & 5.79 & 7.31 & 14.90 & 3677.52 & 52399.01  \\

    \midrule
    \footnotesize{\textbf{Accuracy: 59.69\%}} & & \multicolumn{5}{c}{\textbf{Sparsity}}\\
    \cmidrule{3-7}
        \textbf{Method} & \% trainable & 30\% & 40\% & 50\% & 60\% & 70\% \\
    \midrule
    \glsshort{masklora} & 0.590\% & 59.77\% & 59.33\% & 55.39\% & 52.89\% & 45.42\%  \\
    \gls{ln}-Parameters & 0.004\% & 59.37\% & 58.16\% & 54.74\% & 49.17\% & 40.42\%  \\
    No retraining & 0.000\% & 58.69\% & 56.56\% & 51.14\% & 39.54\% & 33.39\%  \\
    \bottomrule
    
    \\
    \large{\textbf{LLaMA-2-13B}}\\
    \toprule
    \footnotesize{\textbf{Perplexity: 4.57}} & & \multicolumn{5}{c}{\textbf{Sparsity}}\\
    \cmidrule{3-7}
        \textbf{Method} & \% trainable & 30\% & 40\% & 50\% & 60\% & 70\% \\
    \midrule
    \glsshort{masklora} & 0.479\% & 4.75 & 5.00 & 5.50 & 6.56 & 9.09  \\
    \gls{ln}-Parameters & 0.003\% & 4.78 & 5.06 & 5.70 & 7.25 & 11.96  \\
    No retraining & 0.000\% & 4.82 & 5.26 & 6.37 & 11.22 & 275.21  \\

    \midrule
    \footnotesize{\textbf{Accuracy: 62.99\%}} & & \multicolumn{5}{c}{\textbf{Sparsity}}\\
    \cmidrule{3-7}
        \textbf{Method} & \% trainable & 30\% & 40\% & 50\% & 60\% & 70\% \\
    \midrule
    \glsshort{masklora} & 0.479\% & 61.77\% & 60.84\% & 59.23\% & 55.07\% & 47.78\%  \\
    \gls{ln}-Parameters & 0.003\% & 61.65\% & 60.44\% & 58.10\% & 53.04\% & 43.31\%  \\
    No retraining & 0.000\% & 61.99\% & 59.56\% & 52.89\% & 43.90\% & 33.53\%  \\

    \bottomrule
    
    \end{tabular}
    \end{table}

\newpage
\FloatBarrier
\subsection{Closing the Gap: Low-Rank Adaptation}
\FloatBarrier

\begin{table}[h]
    \caption{OPT-2.7B/6.7B/13B/30B: Perplexity comparison of all \gls{lora}-variants retraining magnitude pruned models, both in the unstructured pruning setting (50\% sparsity), as well as for the semi-structured 2:4 and 4:8 sparsities. The second column indicates whether the adapters can be merged into the original weights without destroying the sparsity. We report the mean perplexity over several seeds and omit the standard deviation for the sake of clarity.\\ }
    \centering
    \begin{tabular}{lcc cccc}
    \toprule
     & & & \multicolumn{4}{c}{\textbf{OPT}}\\
    \cmidrule{4-7}
        \textbf{Method} & Mergeable & Sparsity & 2.7B & 6.7B & 13B & 30B \\
    \midrule
    Baseline & -- & 0\% & 12.47 & 10.86 & 10.12 & 9.55\\
    \midrule
    \glsentryshort{lora} & \xmark & 50\% & 15.62 & 13.76 & 12.66 & 11.67  \\
    \glsentryshort{lora-prune} & \cmark & 50\% & 17.86 & 14.97 & 20.87 & 13.37  \\
    \glsentryshort{multlora} & \cmark & 50\% & 16.14 & 14.19 & 13.50 & 12.42  \\
    \glsentryshort{masklora} & \cmark & 50\% & 15.77 & 14.23 & 13.79 & 11.80  \\
    \midrule
    \glsentryshort{lora} & \xmark & 2:4 & 18.14 & 14.55 & 13.20 & 11.98  \\
    \glsentryshort{lora-prune} & \cmark & 2:4 & 20.86 & 15.76 & 15.01 & 13.23  \\
    \glsentryshort{multlora} & \cmark & 2:4 & 18.67 & 14.96 & 13.63 & 12.27  \\
    \glsentryshort{masklora} & \cmark & 2:4 & 18.41 & 14.74 & 13.36 & 12.23  \\
    \midrule
    \glsentryshort{lora} & \xmark & 4:8 & 16.25 & 13.89 & 12.70 & 11.87  \\
    \glsentryshort{lora-prune} & \cmark & 4:8 & 17.70 & 14.80 & 14.25 & 12.87  \\
    \glsentryshort{multlora} & \cmark & 4:8 & 16.58 & 14.49 & 13.12 & 12.29  \\
    \glsentryshort{masklora} & \cmark & 4:8 & 16.46 & 14.08 & 12.94 & 12.05  \\
    \bottomrule
    \end{tabular}
    \end{table}

\begin{table}[h]
    \caption{OPT-2.7B/6.7B/13B/30B: EleutherAI zero-shot accuracy comparison of all \gls{lora}-variants retraining magnitude pruned models, both in the unstructured pruning setting (50\% sparsity), as well as for the semi-structured 2:4 and 4:8 sparsities. The second column indicates whether the adapters can be merged into the original weights without destroying the sparsity. We report the mean accuracy over several seeds and omit the standard deviation for the sake of clarity.\\ }
    \centering
    \begin{tabular}{lcc cccc}
    \toprule
     & & & \multicolumn{4}{c}{\textbf{OPT}}\\
    \cmidrule{4-7}
        \textbf{Method} & Mergeable & Sparsity & 2.7B & 6.7B & 13B & 30B \\
    \midrule
    Baseline & -- & 0\% & 47.81\% & 51.55\% & 52.60\% & 55.07\%\\
    \midrule
    \glsentryshort{lora} & \xmark & 50\% & 46.22\% & 50.16\% & 50.88\% & 52.65\%  \\
    \glsentryshort{lora-prune} & \cmark & 50\% & 45.33\% & 49.61\% & 47.40\% & 50.86\%  \\
    \glsentryshort{multlora} & \cmark & 50\% & 46.03\% & 49.80\% & 50.50\% & 52.31\%  \\
    \glsentryshort{masklora} & \cmark & 50\% & 46.23\% & 50.11\% & 49.65\% & 52.97\%  \\
    \midrule
    \glsentryshort{lora} & \xmark & 2:4 & 44.48\% & 48.68\% & 49.93\% & 49.67\%  \\
    \glsentryshort{lora-prune} & \cmark & 2:4 & 43.84\% & 48.72\% & 48.63\% & 50.09\%  \\
    \glsentryshort{multlora} & \cmark & 2:4 & 44.31\% & 48.20\% & 49.73\% & 49.77\%  \\
    \glsentryshort{masklora} & \cmark & 2:4 & 44.13\% & 48.86\% & 50.00\% & 49.97\%  \\
    \midrule
    \glsentryshort{lora} & \xmark & 4:8 & 45.60\% & 49.67\% & 50.49\% & 51.01\%  \\
    \glsentryshort{lora-prune} & \cmark & 4:8 & 45.17\% & 49.08\% & 48.96\% & 49.52\%  \\
    \glsentryshort{multlora} & \cmark & 4:8 & 45.50\% & 49.45\% & 50.11\% & 50.88\%  \\
    \glsentryshort{masklora} & \cmark & 4:8 & 45.37\% & 49.56\% & 50.66\% & 50.90\%  \\
    \bottomrule
    \end{tabular}
    \end{table}

    \begin{table}[h]
        \caption{LLaMA-2-7B/13B and Mixtral-8x7B: Perplexity comparison of all \gls{lora}-variants retraining magnitude pruned models, both in the unstructured pruning setting (50\% sparsity), as well as for the semi-structured 2:4 and 4:8 sparsities. The second column indicates whether the adapters can be merged into the original weights without destroying the sparsity. We report the mean perplexity over several seeds and omit the standard deviation for the sake of clarity.\\ }
        \centering
        \begin{tabular}{lcc cc | c}
        \toprule
         & & & \multicolumn{2}{c}{\textbf{LLaMA-2}} & \multicolumn{1}{c}{\textbf{Mixtral}}\\
        \cmidrule{4-5} \cmidrule{6-6}
            \textbf{Method} & Mergeable & Sparsity & 7B & 13B & 8x7B \\
        \midrule
        Baseline & -- & 0\% & 5.11 & 4.57 & 3.58 \\
        \midrule
        \glsentryshort{lora} & \xmark & 50\% & 6.42 & 5.46 & 4.76  \\
        \glsentryshort{lora-prune} & \cmark & 50\% & 7.02 & 5.60 & 5.23  \\
        \glsentryshort{multlora} & \cmark & 50\% & 6.70 & 5.54 & 4.90  \\
        \glsentryshort{masklora} & \cmark & 50\% & 6.46 & 5.48 & 4.82  \\
        \midrule
        \glsentryshort{lora} & \xmark & 2:4 & 7.69 & 6.37 & 5.60  \\
        \glsentryshort{lora-prune} & \cmark & 2:4 & 8.25 & 6.56 & 6.18  \\
        \glsentryshort{multlora} & \cmark & 2:4 & 7.99 & 6.48 & 5.81  \\
        \glsentryshort{masklora} & \cmark & 2:4 & 7.72 & 6.38 & 5.65  \\
        \midrule
        \glsentryshort{lora} & \xmark & 4:8 & 6.99 & 5.88 & 5.18  \\
        \glsentryshort{lora-prune} & \cmark & 4:8 & 7.47 & 6.04 & 5.66  \\
        \glsentryshort{multlora} & \cmark & 4:8 & 7.24 & 6.00 & 5.34  \\
        \glsentryshort{masklora} & \cmark & 4:8 & 7.01 & 5.90 & 5.24  \\
        \bottomrule
        \end{tabular}
        \end{table}

        \begin{table}[h]
            \caption{LLaMA-2-7B/13B and Mixtral-8x7B: EleutherAI zero-shot accuracy comparison of all \gls{lora}-variants retraining magnitude pruned models, both in the unstructured pruning setting (50\% sparsity), as well as for the semi-structured 2:4 and 4:8 sparsities. The second column indicates whether the adapters can be merged into the original weights without destroying the sparsity. We report the mean accuracy over several seeds and omit the standard deviation for the sake of clarity.\\ }
            \centering
            \begin{tabular}{lcc cc | c}
            \toprule
             & & & \multicolumn{2}{c}{\textbf{LLaMA-2}} & \multicolumn{1}{c}{\textbf{Mixtral}}\\
            \cmidrule{4-5} \cmidrule{6-6}
                \textbf{Method} & Mergeable & Sparsity & 7B & 13B & 8x7B \\
            \midrule
            Baseline & -- & 0\% & 59.69\% & 62.99\% & 67.70\%\\
            \midrule
            \glsentryshort{lora} & \xmark & 50\% & 56.14\% & 59.21\% & 65.33\%  \\
            \glsentryshort{lora-prune} & \cmark & 50\% & 55.98\% & 58.56\% & 64.17\%  \\
            \glsentryshort{multlora} & \cmark & 50\% & 55.79\% & 59.45\% & 64.84\%  \\
            \glsentryshort{masklora} & \cmark & 50\% & 55.67\% & 59.44\%  & 64.84\%  \\
            \midrule
            \glsentryshort{lora} & \xmark & 2:4 & 51.32\% & 55.68\% & 60.69\%  \\
            \glsentryshort{lora-prune} & \cmark & 2:4 & 51.17\% & 54.56\% & 59.25\%  \\
            \glsentryshort{multlora} & \cmark & 2:4 & 51.15\% & 55.24\% & 60.84\%  \\
            \glsentryshort{masklora} & \cmark & 2:4 & 51.41\% & 55.73\% & 61.01\%  \\
            \midrule
            \glsentryshort{lora} & \xmark & 4:8 & 54.13\% & 58.30\% & 62.78\%  \\
            \glsentryshort{lora-prune} & \cmark & 4:8 & 53.34\% & 57.36\% & 60.72\%  \\
            \glsentryshort{multlora} & \cmark & 4:8 & 53.50\% & 57.07\% & 62.50\%  \\
            \glsentryshort{masklora} & \cmark & 4:8 & 54.29\% & 57.99\% & 62.72\%  \\
            \bottomrule
            \end{tabular}
            \end{table}

    \begin{table}[h]
        \caption{OPT-2.7B/6.7B/13B/30B: Perplexity and EleutherAI zero-shot accuracy comparison of all \gls{lora}-variants retraining magnitude pruned models. We display unstructured pruning sparsities between 40\% and 70\%. We report the mean accuracy over several seeds and omit the standard deviation for the sake of clarity.\\ }

        \centering
        \begin{tabular}{l cccc | cccc}
        \multicolumn{1}{c}{\large{\textbf{OPT-2.7B}}} & \multicolumn{4}{c}{\large{\textbf{Perplexity}}} & \multicolumn{4}{c}{\large{\textbf{Accuracy}}} \\
        \toprule
        \textbf{Method} & \multicolumn{4}{c}{\textbf{Sparsity}} & \multicolumn{4}{c}{\textbf{Sparsity}} \\
        \cmidrule{2-5} \cmidrule{6-9}
         & 40\% & 50\% & 60\% & 70\% & 40\% & 50\% & 60\% & 70\% \\
        \midrule
        \glsentryshort{lora} & 14.26 & 15.63 & 18.20 & 25.38  & 46.17\% & 45.85\% & 43.84\% & 41.05\%  \\
        \glsentryshort{lora-prune} & 14.75 & 17.84 & 27.78 & 92.81  & 46.50\% & 45.37\% & 42.00\% & 38.76\%  \\
        \glsentryshort{multlora} & 14.46 & 16.15 & 19.25 & 25.61  & 45.82\% & 45.95\% & 44.28\% & 41.12\%  \\
        \glsentryshort{masklora} & 14.38 & 15.91 & 18.66 & 25.46  & 46.21\% & 46.12\% & 44.44\% & 41.41\%  \\
        \bottomrule
        \\

        \multicolumn{1}{c}{\large{\textbf{OPT-6.7B}}} & \multicolumn{4}{c}{\large{\textbf{Perplexity}}} & \multicolumn{4}{c}{\large{\textbf{Accuracy}}} \\
        \toprule
        \textbf{Method} & \multicolumn{4}{c}{\textbf{Sparsity}} & \multicolumn{4}{c}{\textbf{Sparsity}} \\
        \cmidrule{2-5} \cmidrule{6-9}
         & 40\% & 50\% & 60\% & 70\% & 40\% & 50\% & 60\% & 70\% \\
        \midrule
        \glsentryshort{lora} & 12.52 & 13.74 & 15.97 & 20.31  & 50.70\% & 50.09\% & 48.76\% & 46.13\%  \\
        \glsentryshort{lora-prune} & 12.97 & 14.97 & 20.39 & 60.68  & 50.69\% & 49.67\% & 47.58\% & 43.08\%  \\
        \glsentryshort{multlora} & 12.74 & 14.12 & 17.08 & 20.93  & 50.77\% & 50.02\% & 48.64\% & 45.28\%  \\
        \glsentryshort{masklora} & 12.57 & 14.25 & 16.76 & 20.83  & 50.75\% & 50.13\% & 48.93\% & 45.93\%  \\
        \bottomrule
        \\

        \multicolumn{1}{c}{\large{\textbf{OPT-13B}}} & \multicolumn{4}{c}{\large{\textbf{Perplexity}}} & \multicolumn{4}{c}{\large{\textbf{Accuracy}}} \\
        \toprule
        \textbf{Method} & \multicolumn{4}{c}{\textbf{Sparsity}} & \multicolumn{4}{c}{\textbf{Sparsity}} \\
        \cmidrule{2-5} \cmidrule{6-9}
         & 40\% & 50\% & 60\% & 70\% & 40\% & 50\% & 60\% & 70\% \\
        \midrule
        \glsentryshort{lora} & 11.62 & 12.62 & 14.35 & 17.22 & 52.07\% & 50.92\% & 49.12\% & 46.45\%  \\
        \glsentryshort{lora-prune} & 12.27 & 15.35 & 23.97 & 138.56  & 51.07\% & 48.62\% & 45.50\% & 40.86\%  \\
        \glsentryshort{multlora} & 11.79 & 13.47 & 15.12 & 18.62  & 51.89\% & 50.69\% & 48.67\% & 46.21\%  \\
        \glsentryshort{masklora} & 11.74 & 13.80 & 15.11 & 17.61  & 51.76\% & 49.91\% & 48.39\% & 46.48\%  \\
        \bottomrule
        \\

        \multicolumn{1}{c}{\large{\textbf{OPT-30B}}} & \multicolumn{4}{c}{\large{\textbf{Perplexity}}} & \multicolumn{4}{c}{\large{\textbf{Accuracy}}} \\
        \toprule
        \textbf{Method} & \multicolumn{4}{c}{\textbf{Sparsity}} & \multicolumn{4}{c}{\textbf{Sparsity}} \\
        \cmidrule{2-5} \cmidrule{6-9}
         & 40\% & 50\% & 60\% & 70\% & 40\% & 50\% & 60\% & 70\% \\
        \midrule
        \glsentryshort{lora} & 10.99 & 11.63 & 13.24 & 16.34  & 53.22\% & 52.79\% & 50.68\% & 47.41\%  \\
        \glsentryshort{lora-prune} & 11.63 & 13.35 & 21.69 & 229.43 & 52.33\% & 50.53\% & 48.81\% & 40.06\%  \\
        \glsentryshort{multlora} & 11.20 & 12.46 & 13.81 & 17.06  & 52.92\% & 52.37\% & 50.08\% & 47.70\%  \\
        \glsentryshort{masklora} & 11.05 & 11.82 & 13.45 & 17.29  & 53.19\% & 52.92\% & 50.55\% & 47.90\%  \\
        \bottomrule

        \end{tabular}
    \end{table}

    \begin{table}[h]
        \caption{LLaMA-2-7B/13B: Perplexity and EleutherAI zero-shot accuracy comparison of all \gls{lora}-variants retraining magnitude pruned models. We display unstructured pruning sparsities between 40\% and 70\%. We report the mean accuracy over several seeds and omit the standard deviation for the sake of clarity.\\ }

        \centering
        \begin{tabular}{l cccc | cccc}
        \multicolumn{1}{c}{\large{\textbf{LLaMA-2-7B}}} & \multicolumn{4}{c}{\large{\textbf{Perplexity}}} & \multicolumn{4}{c}{\large{\textbf{Accuracy}}} \\
        \toprule
        \textbf{Method} & \multicolumn{4}{c}{\textbf{Sparsity}} & \multicolumn{4}{c}{\textbf{Sparsity}} \\
        \cmidrule{2-5} \cmidrule{6-9}
         & 40\% & 50\% & 60\% & 70\% & 40\% & 50\% & 60\% & 70\% \\
        \midrule
        \glsentryshort{lora} & 5.72 & 6.42 & 7.86 & 11.54  & 58.69\% & 56.12\% & 53.58\% & 46.80\%  \\
        \glsentryshort{lora-prune} & 5.99 & 7.07 & 9.47 & 21.69  & 59.00\% & 55.93\% & 51.07\% & 42.55\%  \\
        \glsentryshort{multlora} & 5.84 & 6.67 & 8.29 & 12.56  & 58.38\% & 55.85\% & 51.70\% & 45.03\%  \\
        \glsentryshort{masklora} & 5.77 & 6.46 & 7.98 & 11.56  & 59.34\% & 55.67\% & 52.50\% & 46.27\%  \\
        \bottomrule
        \\

        \multicolumn{1}{c}{\large{\textbf{LLaMA-2-13B}}} & \multicolumn{4}{c}{\large{\textbf{Perplexity}}} & \multicolumn{4}{c}{\large{\textbf{Accuracy}}} \\
        \toprule
        \textbf{Method} & \multicolumn{4}{c}{\textbf{Sparsity}} & \multicolumn{4}{c}{\textbf{Sparsity}} \\
        \cmidrule{2-5} \cmidrule{6-9}
         & 40\% & 50\% & 60\% & 70\% & 40\% & 50\% & 60\% & 70\% \\
        \midrule
        \glsentryshort{lora} & 4.99 & 5.46 & 6.47 & 8.97  & 60.43\% & 59.16\% & 55.41\% & 49.25\%  \\
        \glsentryshort{lora-prune} & 5.05 & 5.61 & 6.96 & 11.10  & 60.68\% & 58.58\% & 53.54\% & 44.42\%  \\
        \glsentryshort{multlora} & 5.01 & 5.54 & 6.65 & 9.48  & 60.28\% & 59.18\% & 55.50\% & 49.14\%  \\
        \glsentryshort{masklora} & 4.99 & 5.48 & 6.51 & 9.04  & 60.69\% & 59.34\% & 55.28\% & 49.40\%  \\
        \bottomrule

        \end{tabular}
    \end{table}

\clearpage

\FloatBarrier
\subsection{Efficient Layer-wise Reconstruction}\label{subsec:app_reconstruction}
\FloatBarrier
\begin{table}[h]
    \caption{OPT-2.7B/6.7B/13B/30B: Perplexity comparison of magnitude pruning, Wanda and SparseGPT with and without \glsentryshort{masklora} when reconstructing instead of retraining, both in the unstructured pruning setting (50\% sparsity), as well as for the semi-structured 2:4 and 4:8 sparsities. We report the mean perplexity over several seeds and omit the standard deviation for the sake of clarity.\\ }
    \label{tab:perplexity_reconstruction_comparison}
    \centering
    \begin{tabular}{lcc cccc}
    \toprule
     & & & \multicolumn{4}{c}{\textbf{OPT}}\\
    \cmidrule{4-7}
        \textbf{Method} & Reconstruction & Sparsity & 2.7B & 6.7B & 13B & 30B \\
    \midrule
    Baseline & -- & 0\% & 12.47 & 10.86 & 10.12 & 9.55\\
    \midrule
    Magnitude & \xmark & 50\% & 265.14 & 968.69 & 11558.65 & 168.06  \\
    Magnitude & \cmark & 50\% & 15.22 & 13.14 & 12.34 & 11.06  \\
    Wanda & \xmark & 50\% & 22.79 & 15.43 & 13.55 & 10.86  \\
    Wanda & \cmark & 50\% & 13.91 & 11.82 & 11.12 & 10.01  \\
    SparseGPT & \xmark & 50\% & 13.50 & 11.58 & 11.21 & 9.79  \\
    SparseGPT & \cmark & 50\% & 13.42 & 11.48 & 10.85 & 9.79  \\
    \midrule
    Magnitude & \xmark & 2:4 & 1153.15 & 264.11 & 484.84 & 1979.66  \\
    Magnitude & \cmark & 2:4 & 18.16 & 14.90 & 13.07 & 39.12  \\
    Wanda & \xmark & 2:4 & 21.38 & 16.04 & 15.74 & 13.26  \\
    Wanda & \cmark & 2:4 & 17.54 & 14.43 & 12.59 & 11.26  \\
    SparseGPT & \xmark & 2:4 & 17.28 & 14.27 & 13.00 & 10.95  \\
    SparseGPT & \cmark & 2:4 & 15.93 & 13.32 & 11.93 & 10.49  \\
    \midrule
    Magnitude & \xmark & 4:8 & 166.91 & 196.18 & 450.02 & 563.77  \\
    Magnitude & \cmark & 4:8 & 15.95 & 13.47 & 12.16 & 14.34  \\
    Wanda & \xmark & 4:8 & 16.90 & 13.64 & 13.47 & 10.88  \\
    Wanda & \cmark & 4:8 & 15.38 & 12.75 & 11.74 & 10.49  \\
    SparseGPT & \xmark & 4:8 & 15.06 & 12.60 & 11.79 & 10.31  \\
    SparseGPT & \cmark & 4:8 & 14.51 & 12.16 & 11.18 & 10.10  \\
    \bottomrule
    \end{tabular}
    \end{table}

    \begin{table}[h]
        \caption{OPT-2.7B/6.7B/13B/30B: Zero-shot accuracy comparison on the EleutherAI evaluation set. We report magnitude pruning, Wanda and SparseGPT with and without \glsentryshort{masklora} when reconstructing instead of retraining, both in the unstructured pruning setting (50\% sparsity), as well as for the semi-structured 2:4 and 4:8 sparsities. We report the mean accuracy over several seeds and omit the standard deviation for the sake of clarity.\\ }
        \label{tab:accuracy_reconstruction_comparison_including_FT}
        \centering
        \begin{tabular}{lcc cccc}
        \toprule
         & & & \multicolumn{4}{c}{\textbf{OPT}}\\
        \cmidrule{4-7}
            \textbf{Method} & Reconstruction & Sparsity & 2.7B & 6.7B & 13B & 30B \\
        \midrule
        Baseline & -- & 0\% & 47.81\% & 51.55\% & 52.60\% & 55.07\%\\
        \midrule
        Magnitude & \xmark & 50\% & 40.07\% & 35.54\% & 33.80\% & 36.39\%  \\
        Magnitude & \cmark & 50\% & 45.14\% & 48.99\% & 50.41\% & 51.81\%  \\
        Wanda & \xmark & 50\% & 42.63\% & 47.14\% & 50.34\% & 53.15\%  \\
        Wanda & \cmark & 50\% & 46.47\% & 49.81\% & 51.65\% & 54.00\%  \\
        SparseGPT & \xmark & 50\% & 46.53\% & 50.26\% & 51.93\% & 54.01\%  \\
        SparseGPT & \cmark & 50\% & 46.62\% & 50.42\% & 51.92\% & 54.33\%  \\
        \midrule
        Magnitude & \xmark & 2:4 & 35.94\% & 36.38\% & 36.65\% & 34.92\%  \\
        Magnitude & \cmark & 2:4 & 43.54\% & 47.70\% & 49.16\% & 47.74\%  \\
        Wanda & \xmark & 2:4 & 42.82\% & 46.12\% & 47.70\% & 49.69\%  \\
        Wanda & \cmark & 2:4 & 43.69\% & 47.46\% & 48.56\% & 50.93\%  \\
        SparseGPT & \xmark & 2:4 & 44.12\% & 47.28\% & 48.84\% & 51.17\%  \\
        SparseGPT & \cmark & 2:4 & 44.67\% & 48.39\% & 50.38\% & 52.33\%  \\
        \midrule
        Magnitude & \xmark & 4:8 & 36.95\% & 36.90\% & 36.13\% & 36.81\%  \\
        Magnitude & \cmark & 4:8 & 44.59\% & 48.51\% & 50.19\% & 50.52\%  \\
        Wanda & \xmark & 4:8 & 44.02\% & 47.45\% & 49.10\% & 51.20\%  \\
        Wanda & \cmark & 4:8 & 44.69\% & 48.64\% & 50.21\% & 51.83\%  \\
        SparseGPT & \xmark & 4:8 & 44.98\% & 48.33\% & 50.12\% & 52.28\%  \\
        SparseGPT & \cmark & 4:8 & 45.99\% & 49.48\% & 50.96\% & 52.92\%  \\
        \bottomrule
        \end{tabular}
        \end{table}

        \begin{table}[h]
            \caption{LLaMA-2-7B/13B and Mistral-7B: Perplexity comparison of magnitude pruning, Wanda and SparseGPT with and without \glsentryshort{masklora} when reconstructing instead of retraining, both in the unstructured pruning setting (50\% sparsity), as well as for the semi-structured 2:4 and 4:8 sparsities. We report the mean perplexity over several seeds and omit the standard deviation for the sake of clarity.\\ }
            \centering
            \begin{tabular}{lcc cc | c}
            \toprule
             & & & \multicolumn{2}{c}{\textbf{LLaMA-2}} & \multicolumn{1}{c}{\textbf{Mistral}}\\
            \cmidrule{4-5} \cmidrule{6-6}
                \textbf{Method} & Reconstruction & Sparsity & 7B & 13B & 7B \\
            \midrule
            Baseline & -- & 0\% & 5.11 & 4.57 & 4.90 \\
            \midrule
            Magnitude & \xmark & 50\% & 14.90 & 6.37 & 7.42  \\
            Magnitude & \cmark & 50\% & 9.54 & 5.43 & 6.17  \\
            Wanda & \xmark & 50\% & 7.38 & 5.93 & 6.05  \\
            Wanda & \cmark & 50\% & 6.16 & 5.30 & 5.83  \\
            SparseGPT & \xmark & 50\% & 6.53 & 5.64 & 6.00  \\
            SparseGPT & \cmark & 50\% & 6.18 & 5.37 & 5.81  \\
            \midrule
            Magnitude & \xmark & 2:4 & 54.39 & 8.32 & 14.15  \\
            Magnitude & \cmark & 2:4 & 21.67 & 6.72 & 10.44  \\
            Wanda & \xmark & 2:4 & 11.35 & 8.36 & 11.55  \\
            Wanda & \cmark & 2:4 & 8.14 & 6.59 & 8.46  \\
            SparseGPT & \xmark & 2:4 & 10.23 & 8.30 & 9.57  \\
            SparseGPT & \cmark & 2:4 & 8.07 & 6.70 & 7.83  \\
            \midrule
            Magnitude & \xmark & 4:8 & 16.53 & 6.76 & 9.10  \\
            Magnitude & \cmark & 4:8 & 11.48 & 5.99 & 7.72  \\
            Wanda & \xmark & 4:8 & 8.07 & 6.55 & 7.82  \\
            Wanda & \cmark & 4:8 & 6.97 & 5.88 & 6.80  \\
            SparseGPT & \xmark & 4:8 & 8.00 & 6.58 & 7.41  \\
            SparseGPT & \cmark & 4:8 & 6.98 & 5.95 & 6.66  \\
            \bottomrule
            \end{tabular}
            \end{table}

        \begin{table}[h]
            \caption{LLaMA-2-7B/13B and Mistral-7B: Zero-shot accuracy comparison on the EleutherAI evaluation set. We report magnitude pruning, Wanda and SparseGPT with and without \glsentryshort{masklora} when reconstructing instead of retraining, both in the unstructured pruning setting (50\% sparsity), as well as for the semi-structured 2:4 and 4:8 sparsities. We report the mean accuracy over several seeds and omit the standard deviation for the sake of clarity.\\ }
                \centering
                \begin{tabular}{lcc cc | c}
                \toprule
                 & & & \multicolumn{2}{c}{\textbf{LLaMA-2}} & \multicolumn{1}{c}{\textbf{Mistral}}\\
                \cmidrule{4-5} \cmidrule{6-6}
                    \textbf{Method} & Reconstruction & Sparsity & 7B & 13B & 7B \\
                \midrule
                Baseline & -- & 0\% & 59.69\% & 62.99\% & 64.36\%\\
                \midrule
                Magnitude & \xmark & 50\% & 51.15\% & 52.88\% & 55.86\%  \\
                Magnitude & \cmark & 50\% & 55.08\% & 59.55\% & 59.18\%  \\
                Wanda & \xmark & 50\% & 54.89\% & 58.27\% & 59.62\%  \\
                Wanda & \cmark & 50\% & 55.87\% & 60.46\% & 59.52\%  \\
                SparseGPT & \xmark & 50\% & 55.86\% & 60.98\% & 59.86\%  \\
                SparseGPT & \cmark & 50\% & 56.14\% & 61.26\% & 60.24\%  \\
                \midrule
                Magnitude & \xmark & 2:4 & 47.48\% & 49.77\% & 49.69\%  \\
                Magnitude & \cmark & 2:4 & 49.44\% & 54.43\% & 52.04\%  \\
                Wanda & \xmark & 2:4 & 48.74\% & 53.20\% & 50.09\%  \\
                Wanda & \cmark & 2:4 & 50.87\% & 55.37\% & 51.68\%  \\
                SparseGPT & \xmark & 2:4 & 50.83\% & 55.48\% & 53.01\%  \\
                SparseGPT & \cmark & 2:4 & 52.08\% & 56.96\% & 55.06\%  \\
                \midrule
                Magnitude & \xmark & 4:8 & 50.68\% & 52.72\% & 54.32\%  \\
                Magnitude & \cmark & 4:8 & 52.44\% & 58.03\% & 56.13\%  \\
                Wanda & \xmark & 4:8 & 52.55\% & 58.45\% & 55.30\%  \\
                Wanda & \cmark & 4:8 & 53.05\% & 58.99\% & 55.68\%  \\
                SparseGPT & \xmark & 4:8 & 53.60\% & 58.17\% & 57.42\%  \\
                SparseGPT & \cmark & 4:8 & 54.42\% & 59.38\% & 58.55\%  \\
                \bottomrule
                \end{tabular}
                \end{table}

        \begin{table}[h]
            \caption{OPT-6.7B/13B: Zero-shot accuracy of \glsshort{masklora} and full \gls{ft} with 40\%-70\% of the parameters pruned in the reconstruction setting. The first column lists the method. Full \gls{ft} represents the baseline of reconstructing all parameters and is only possible in the reconstruction setting for models not bigger than OPT-13B due to memory constraints. The next columns display the average zero-shot accuracy (higher is better) across multiple seeds, with standard deviations excluded for clarity.\\}
            \label{tab:fullft_reconstruction_accuracy_oneshot_comparison}
            \centering
            \begin{tabular}{l cccc | cccc}
            \multicolumn{1}{c}{} & \multicolumn{4}{c}{\large{\textbf{OPT-6.7B}}} & \multicolumn{4}{c}{\large{\textbf{OPT-13B}}} \\
            \toprule
            \textbf{Method} & \multicolumn{4}{c}{\textbf{Sparsity}} & \multicolumn{4}{c}{\textbf{Sparsity}} \\
            \cmidrule{2-5} \cmidrule{6-9}
             & 40\% & 50\% & 60\% & 70\% & 40\% & 50\% & 60\% & 70\% \\
            \midrule
            Full \gls{ft} & 50.25\% & 48.76\% & 46.95\% & 38.57\% & 51.92\% & 50.78\% & 46.44\% & 41.04\% \\
            \glsshort{masklora} & 50.37\% & 49.00\% & 47.43\% & 41.73\% & 51.65\% & 50.45\% & 48.18\% & 42.13\% \\
            \bottomrule
            \end{tabular}
        \end{table}

\clearpage

\FloatBarrier
\section{Results for convolutional architectures}\label{app:conv_results}
\FloatBarrier

\begin{figure}[h]
    \centering
    \includegraphics[width=0.5\columnwidth]{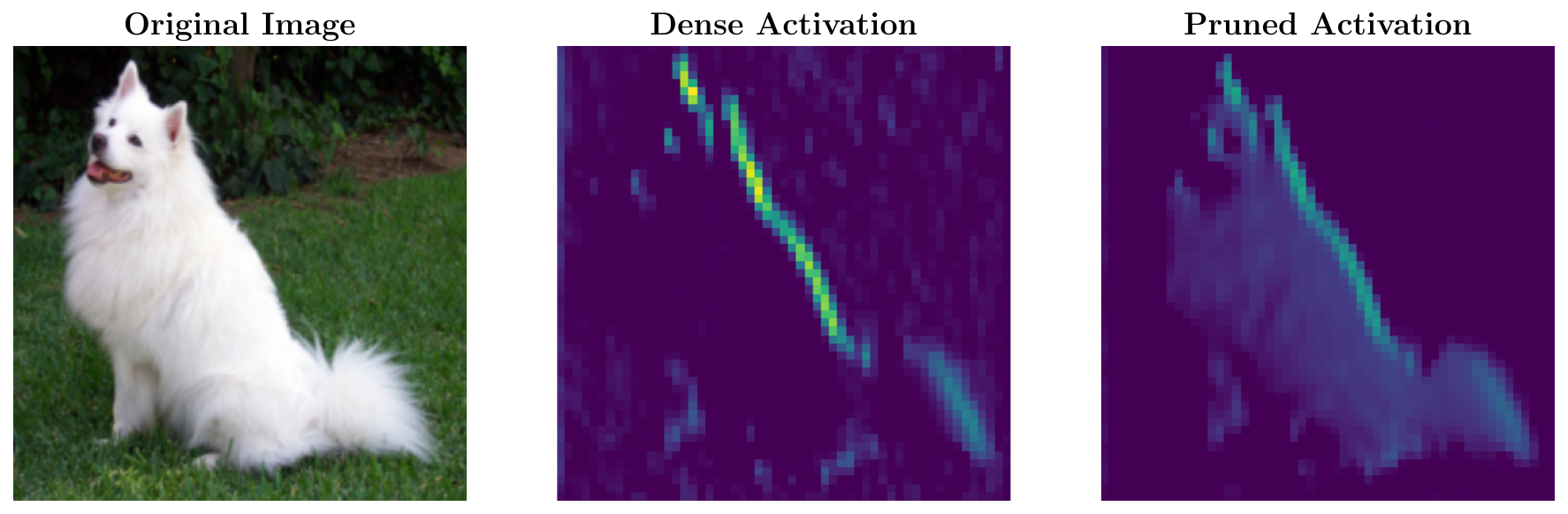}

    \caption{Features produced by a single filter from the first convolutional layer of \emph{AlexNet} \citep{Krizhevsky2012}. From left to right: original image, output from a pretrained model, and output from the magnitude-pruned version of the same model.}
    \label{fig:dog_features}
\end{figure}

\autoref{fig:dog_features} illustrates our intuition by depicting a dog (left) and the features produced by a single filter from the first convolutional layer of a pretrained network (middle) and its pruned version (right). The middle image demonstrates the pretrained network's capability to capture distinct boundary features, especially the dog's defining back and ears. Conversely, the pruned network still emphasizes the dog's back, albeit with reduced intensity and in favor of its overall form, likely influenced by the stark contrast between the white dog and the green grass. While pruning diminishes the feature quality, it does not completely eradicate it.

We investigate the following approaches:
\begin{itemize}

	\item \textbf{BN-Recalibrate}: \citet{Li2020d} identified that recalibrating the \gls{bn} statistics after pruning enhances generalization. This approach entails a one-time evaluation on the training dataset, neither requiring backpropagation nor altering the training set performance.
	\item \textbf{Biases}: Similarly as for \gls{llms}, we only retrain the network's biases.
	\item \textbf{\gls{bn}-Parameters}: Beyond statistics, \gls{bn} layers also include trainable scaling and bias parameters. Similar to \gls{ln}-parameters for \gls{llms}, we retrain these parameters.
	\item \textbf{Linear Head}: We only retrain the linear head.
\end{itemize}

\begin{figure}
    \centering
    \includegraphics[width=0.4\columnwidth]{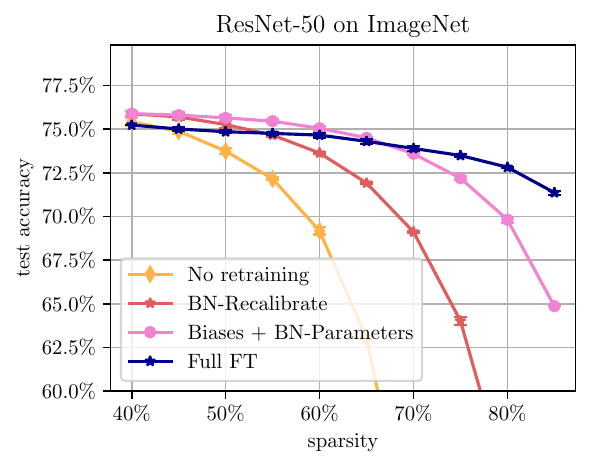}

    \caption{ResNet-50 on ImageNet: Test accuracy across sparsity levels for One Shot pruning with one retraining epoch.}
    \label{fig:imagenet_oneshot_comparison}
\end{figure}

\begin{figure}
        \centering
        \includegraphics[width=0.4\columnwidth]{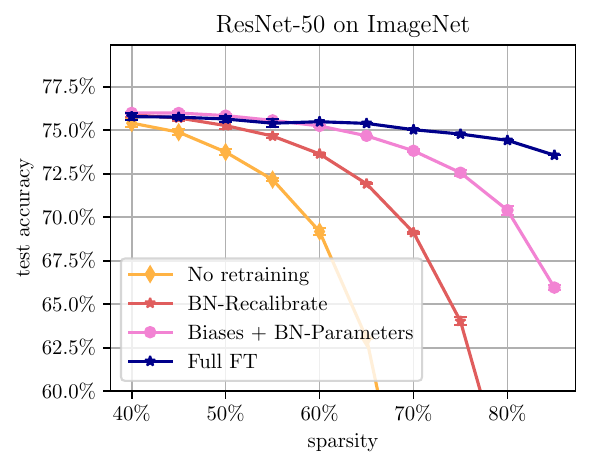}

        \caption{ResNet-50 on ImageNet: Test accuracy across sparsity levels for One Shot pruning with five retraining epochs.}
        \label{appfig:imagenet_oneshot_comparison_5}
\end{figure}

 \autoref{fig:imagenet_oneshot_comparison} and \autoref{appfig:imagenet_oneshot_comparison_5} compare the different approaches on ResNet-50 on ImageNet for one and five retraining epochs, respectively. Since retraining biases and \gls{bn}-parameters individually yield marginal improvements, we only include their combination. We did not find it beneficial to retrain the linear head and have not included it in the plots. Here, \gls{bn}-Recalibrate does not backpropagate any parameters. Combining biases and \gls{bn}-parameters corresponds to retraining roughly 0.21\% of the parameters only. Yet, these approaches are able to recover much of the performance lost due to pruning (comparing to no retraining). At the high sparsity regime, full retraining prevails.

 \newpage

\newpage

\clearpage

\FloatBarrier
\section{Ablation studies}\label{app:ablations}
\FloatBarrier
\subsection{Ablation: Dissecting the impact of parameter groups for retraining}
\FloatBarrier

In \autoref{apptab:ablation_impact_of_groups_50} and \autoref{apptab:ablation_impact_of_groups_70}, we analyze the effect of different parameter groups when retraining OPT-13B for 1000 iterations, magnitude-pruned to 50\% and 70\% sparsity, respectively. We tuned the learning rate and report the best mean perplexity across multiple random seeds, omitting the standard deviation for clarity. The last two columns show the proportion of trainable parameters and final test perplexity, respectively. The first five columns indicate whether individual parameter groups are active (\cmark) or inactive (\xmark), specifically biases, \gls{ln}-parameters, the linear head, the embedding layer, and \glsshort{masklora} parameters.

\begin{table}[h]
\caption{OPT-13B: The subtables report parameter-efficient retraining for 50\% unstructured pruning. The first five columns specify whether certain model parameter subgroups are inactive (\xmark) or active (\cmark), representing the training of biases, \gls{ln} parameters, the linear head, the embedding layer, and using \gls{masklora}, respectively. The penultimate column presents the percentage of trainable parameters in each configuration. The final column displays the mean perplexity across multiple seeds, omitting the standard deviation for clarity.\\ }
\label{apptab:ablation_impact_of_groups_50}
\centering
\begin{tabular}{ccccc|cc}
\large{50\%}\\
\toprule
    Biases & \gls{ln} & Head & Embedding & \glsshort{masklora} & \% trainable & Perplexity \\
\midrule
\xmark & \xmark & \xmark & \xmark & \xmark &  0.00\% & 11558.65  \\
\cmark & \xmark & \xmark & \xmark & \xmark &  0.02\% & 13.29  \\
\xmark & \cmark & \xmark & \xmark & \xmark &  0.01\% & 13.59  \\
\xmark & \xmark & \cmark & \xmark & \xmark &  0.01\% & 2743.37  \\
\xmark & \xmark & \xmark & \cmark & \xmark &  0.01\% & 9450.35  \\
\xmark & \xmark & \xmark & \xmark & \cmark &  0.46\% & 13.59  \\
\cmark & \cmark & \xmark & \xmark & \xmark &  0.02\% & 13.31  \\
\cmark & \xmark & \cmark & \xmark & \xmark &  0.02\% & 13.42  \\
\cmark & \xmark & \xmark & \cmark & \xmark &  0.02\% & 17.71  \\
\cmark & \xmark & \xmark & \xmark & \cmark &  0.47\% & 13.88  \\
\xmark & \cmark & \cmark & \xmark & \xmark &  0.01\% & 13.81  \\
\xmark & \cmark & \xmark & \cmark & \xmark &  0.01\% & 26.51  \\
\xmark & \cmark & \xmark & \xmark & \cmark &  0.46\% & 13.67  \\
\xmark & \xmark & \cmark & \cmark & \xmark &  0.01\% & 2470.32  \\
\xmark & \xmark & \cmark & \xmark & \cmark &  0.46\% & 13.62  \\
\xmark & \xmark & \xmark & \cmark & \cmark &  0.46\% & 16.38  \\
\cmark & \cmark & \cmark & \xmark & \xmark &  0.03\% & 13.44  \\
\cmark & \cmark & \xmark & \cmark & \xmark &  0.03\% & 16.84  \\
\cmark & \cmark & \xmark & \xmark & \cmark &  0.48\% & 13.44  \\
\cmark & \xmark & \cmark & \cmark & \xmark &  0.03\% & 18.39  \\
\cmark & \xmark & \cmark & \xmark & \cmark &  0.48\% & 13.79  \\
\cmark & \xmark & \xmark & \cmark & \cmark &  0.48\% & 15.46  \\
\xmark & \cmark & \cmark & \cmark & \xmark &  0.02\% & 25.21  \\
\xmark & \cmark & \cmark & \xmark & \cmark &  0.47\% & 13.57  \\
\xmark & \cmark & \xmark & \cmark & \cmark &  0.47\% & 16.54  \\
\xmark & \xmark & \cmark & \cmark & \cmark &  0.47\% & 16.14  \\
\cmark & \cmark & \cmark & \cmark & \xmark &  0.03\% & 17.10  \\
\cmark & \cmark & \cmark & \xmark & \cmark &  0.48\% & 13.69  \\
\cmark & \cmark & \xmark & \cmark & \cmark &  0.48\% & 16.33  \\
\cmark & \xmark & \cmark & \cmark & \cmark &  0.49\% & 16.63  \\
\xmark & \cmark & \cmark & \cmark & \cmark &  0.48\% & 15.60  \\
\cmark & \cmark & \cmark & \cmark & \cmark &  0.49\% & 16.19  \\
\bottomrule

\end{tabular}
\end{table}

\begin{table}[h]
    \caption{OPT-13B: The subtables report parameter-efficient retraining for 70\% unstructured pruning. The first five columns specify whether certain model parameter subgroups are inactive (\xmark) or active (\cmark), representing the training of biases, \gls{ln} parameters, the linear head, the embedding layer, and using \gls{masklora}, respectively. The penultimate column presents the percentage of trainable parameters in each configuration. The final column displays the mean perplexity across seeds, where we omit the standard deviation for clarity.\\ }
    \label{apptab:ablation_impact_of_groups_70}
    \centering
    \begin{tabular}{ccccc|cc}
    \large{70\%}\\
    \toprule
        Biases & \gls{ln} & Head & Embedding & \glsshort{masklora} & \% trainable & Perplexity \\
    \midrule
    \xmark & \xmark & \xmark & \xmark & \xmark &  0.00\% & 289842.47  \\
    \cmark & \xmark & \xmark & \xmark & \xmark &  0.02\% & 20.95  \\
    \xmark & \cmark & \xmark & \xmark & \xmark &  0.01\% & 26.06  \\
    \xmark & \xmark & \cmark & \xmark & \xmark &  0.01\% & 8399.64  \\
    \xmark & \xmark & \xmark & \cmark & \xmark &  0.01\% & 282834.05  \\
    \xmark & \xmark & \xmark & \xmark & \cmark &  0.46\% & 17.61  \\
    \cmark & \cmark & \xmark & \xmark & \xmark &  0.02\% & 20.31  \\
    \cmark & \xmark & \cmark & \xmark & \xmark &  0.02\% & 21.42  \\
    \cmark & \xmark & \xmark & \cmark & \xmark &  0.02\% & 86.28  \\
    \cmark & \xmark & \xmark & \xmark & \cmark &  0.47\% & 18.07  \\
    \xmark & \cmark & \cmark & \xmark & \xmark &  0.01\% & 26.18  \\
    \xmark & \cmark & \xmark & \cmark & \xmark &  0.01\% & 7094.19  \\
    \xmark & \cmark & \xmark & \xmark & \cmark &  0.46\% & 18.32  \\
    \xmark & \xmark & \cmark & \cmark & \xmark &  0.01\% & 9276.99  \\
    \xmark & \xmark & \cmark & \xmark & \cmark &  0.46\% & 18.60  \\
    \xmark & \xmark & \xmark & \cmark & \cmark &  0.46\% & 22.04  \\
    \cmark & \cmark & \cmark & \xmark & \xmark &  0.03\% & 21.07  \\
    \cmark & \cmark & \xmark & \cmark & \xmark &  0.03\% & 63.86  \\
    \cmark & \cmark & \xmark & \xmark & \cmark &  0.48\% & 18.39  \\
    \cmark & \xmark & \cmark & \cmark & \xmark &  0.03\% & 77.54  \\
    \cmark & \xmark & \cmark & \xmark & \cmark &  0.48\% & 17.65  \\
    \cmark & \xmark & \xmark & \cmark & \cmark &  0.48\% & 22.19  \\
    \xmark & \cmark & \cmark & \cmark & \xmark &  0.02\% & 7403.60  \\
    \xmark & \cmark & \cmark & \xmark & \cmark &  0.47\% & 18.38  \\
    \xmark & \cmark & \xmark & \cmark & \cmark &  0.47\% & 22.35  \\
    \xmark & \xmark & \cmark & \cmark & \cmark &  0.47\% & 22.75  \\
    \cmark & \cmark & \cmark & \cmark & \xmark &  0.03\% & 153.02  \\
    \cmark & \cmark & \cmark & \xmark & \cmark &  0.48\% & 18.40  \\
    \cmark & \cmark & \xmark & \cmark & \cmark &  0.48\% & 21.65  \\
    \cmark & \xmark & \cmark & \cmark & \cmark &  0.49\% & 22.32  \\
    \xmark & \cmark & \cmark & \cmark & \cmark &  0.48\% & 22.37  \\
    \cmark & \cmark & \cmark & \cmark & \cmark &  0.49\% & 21.87  \\
    \bottomrule
    
    \end{tabular}
    \end{table}

\newpage

\FloatBarrier
\subsection{Ablation: The high sparsity regime}
\FloatBarrier
\autoref{apptab:perplexity_saliency_comparison_high_sparsity_reconstruction} examines the perplexity results for unstructured pruning on OPT-30B using magnitude pruning, Wanda, and SparseGPT, both with and without \glsshort{masklora} reconstruction. Analogously, \autoref{apptab:perplexity_saliency_comparison_high_sparsity_retraining} shows the result when retraining instead of reconstructing. Both tables focus on higher sparsity levels to explore whether our findings extend beyond the 50\% threshold.
We make the following observations for the high sparsity regime:
\begin{itemize}
    \item All methods benefit significantly from \glsshort{masklora} retraining or reconstruction. See also \autoref{apptab:task_accuracy_comparison_high_sparsity_retraining} for the improvement on the individual tasks.
    \item Retraining is much more effective than reconstruction.
    \item Only SparseGPT is able to maintain reasonable test perplexity at 80\% sparsity.
    \item Magnitude pruning with \glsshort{masklora} reconstruction or retraining often outperforms Wanda without any further reoptimization; surprisingly, this is also often the case when Wanda pruning is followed by retraining or reconstruction as well.
\end{itemize}

\begin{table}[h]
\caption{OPT-30B Reconstruction: Perplexity comparison of magnitude pruning, Wanda and SparseGPT with and without \glsentryshort{masklora} when reconstructing instead of retraining, displaying the high sparsity regime of unstructured pruning. We report the mean perplexity over several seeds and omit the standard deviation for the sake of clarity.\\ }
\label{apptab:perplexity_saliency_comparison_high_sparsity_reconstruction}
\centering
\begin{tabular}{lc cccc}
\toprule
& & \multicolumn{4}{c}{\textbf{Sparsity}}\\
\cmidrule{3-6}
    \textbf{Method} & Reconstruction & 50\% & 60\% & 70\% & 80\% \\
\midrule
Magnitude & \xmark & 168.06 & 11676.00 & 28180.15 & 56381.68  \\
Magnitude & \cmark & 11.09 & 32.27 & 2659.70 & 4717.70  \\
Wanda & \xmark & 10.86 & 16.94 & 14080.80 & 11654.38  \\
Wanda & \cmark & 10.01 & 11.24 & 121.10 & 9321.66  \\
SparseGPT & \xmark & 9.79 & 10.70 & 13.55 & 47.68  \\
SparseGPT & \cmark & 9.79 & 10.48 & 12.59 & 26.09  \\
\bottomrule
\end{tabular}
\end{table}

\begin{table}[h]
    \caption{OPT-30B Retraining: Perplexity comparison of magnitude pruning, Wanda and SparseGPT with and without \glsentryshort{masklora} retraining, displaying the high sparsity regime of unstructured pruning. We report the mean perplexity over several seeds and omit the standard deviation for the sake of clarity.\\ }
    \label{apptab:perplexity_saliency_comparison_high_sparsity_retraining}
    \centering
    \begin{tabular}{lc cccc}
    \toprule
    & & \multicolumn{4}{c}{\textbf{Sparsity}}\\
    \cmidrule{3-6}
        \textbf{Method} & Reconstruction & 50\% & 60\% & 70\% & 80\% \\
    \midrule
    Magnitude & \xmark & 168.06 & 11676.00 & 28180.15 & 56381.68  \\
    Magnitude & \cmark & 11.77 & 13.61 & 16.88 & 47.25  \\
    Wanda & \xmark & 10.86 & 16.94 & 13667.60 & 12370.62  \\
    Wanda & \cmark & 10.30 & 11.58 & 23.57 & 84.06  \\
    SparseGPT & \xmark & 9.77 & 10.73 & 13.58 & 48.12  \\
    SparseGPT & \cmark & 9.77 & 10.45 & 11.92 & 18.92  \\
    \bottomrule
    \end{tabular}
    \end{table}

\begin{table}[h]
    \caption{OPT-30B Retraining: Task performance improvement by retraining with \glsentryshort{masklora} for magnitude pruning, Wanda, and SparseGPT, displaying the high sparsity regime of unstructured pruning. Each entry shows the improvement compared to no retraining, e.g., +0.5\% indicates a 0.5\% performance increase. We report the mean performance over several seeds and omit the standard deviation for the sake of clarity.\\ }
    \label{apptab:task_accuracy_comparison_high_sparsity_retraining}
    \centering
    \begin{tabular}{lc ccccccc|c}
    \toprule
    \textbf{Method} & \textbf{Sparsity} & \multicolumn{8}{c}{\textbf{\boldmath$\Delta$ Task Accuracy}} \\
    \cmidrule{3-10}
    & & \textbf{BoolQ} & \textbf{RTE} & \textbf{HSwag} & \textbf{WinoG} & \textbf{ARC-e} & \textbf{ARC-c} & \textbf{OBQA} & \textbf{Average} \\
    \midrule
    Magnitude & 50\% & +29.50\% & +6.32\% & +18.82\% & +14.76\% & +25.67\% & +11.48\% & +13.80\% & +16.40\%  \\
    Wanda & 50\% & +0.38\% & +5.23\% & +0.94\% & +1.46\% & +0.29\% & +0.68\% & +1.30\% & +1.24\%  \\
    SparseGPT & 50\% & +0.40\% & +4.87\% & +0.41\% & -0.20\% & -0.02\% & +0.43\% & +0.50\% & +0.60\%  \\
    \midrule
    Magnitude & 60\% & +28.07\% & +1.44\% & +22.93\% & +14.64\% & +37.98\% & +12.24\% & +14.40\% & +18.82\%  \\
    Wanda & 60\% & +1.56\% & +4.69\% & +6.05\% & +3.95\% & +6.99\% & +3.28\% & +3.20\% & +3.84\%  \\
    SparseGPT & 60\% & +4.48\% & +0.90\% & +1.57\% & +0.32\% & +0.55\% & +0.55\% & +2.10\% & +0.99\%  \\
    \midrule
    Magnitude & 70\% & +26.41\% & +2.89\% & +19.85\% & +11.88\% & +35.56\% & +9.04\% & +12.60\% & +16.17\%  \\
    Wanda & 70\% & +24.40\% & +1.08\% & +10.10\% & +5.80\% & +26.33\% & +2.90\% & +7.60\% & +10.78\%  \\
    SparseGPT & 70\% & +4.92\% & +3.61\% & +3.65\% & +1.85\% & +3.03\% & +3.07\% & +2.40\% & +2.66\%  \\
    \midrule
    Magnitude & 80\% & +14.57\% & +0.36\% & +10.72\% & +6.91\% & +25.21\% & +2.26\% & +5.30\% & +9.33\%  \\
    Wanda & 80\% & +9.66\% & 0.00\% & +1.92\% & +0.87\% & +9.51\% & 0.00\% & +1.00\% & +2.30\%  \\
    SparseGPT & 80\% & +0.12\% & +2.35\% & +7.81\% & +5.09\% & +10.90\% & +4.14\% & +3.70\% & +4.66\%  \\

    \bottomrule
    \end{tabular}
\end{table}


\end{document}